\documentclass[sigconf]{acmart}
\usepackage{pifont}
\usepackage{graphicx}
\usepackage{subfigure}
\AtBeginDocument{%
  \providecommand\BibTeX{{%
    \normalfont B\kern-0.5em{\scshape i\kern-0.25em b}\kern-0.8em\TeX}}}

\setcopyright{acmcopyright}
\copyrightyear{2024}
\acmYear{2024}
\acmDOI{XXXXXXX.XXXXXXX}

\acmConference[WWW '24]{ACM Web Conference 2024}{May 13--17,
  2024}{Singapore}
\acmPrice{15.00}
\acmISBN{978-1-4503-XXXX-X/18/06}

\begin{document}

\title{Revisiting VAE for Unsupervised Time Series Anomaly Detection: A Frequency Perspective}


\author{Zexin Wang}
% \email{wangzexin@cnic.cn}
\additionalaffiliation{%
  \institution{University of Chinese Academy of Sciences}
 \city{Beijing}
 \country{China}
 }
\affiliation{%
\institution{Computer Network Information Center, Chinese Academy of Sciences}
 \city{Beijing}
 \country{China}
 }

\author{Changhua Pei}
% \email{chpei@cnic.cn}
\authornote{Corresponding author. Email: chpei@cnic.cn}
\affiliation{%
\institution{Computer Network Information Center, Chinese Academy of Sciences}
 \city{Beijing}
 \country{China}
 }

\author{Minghua Ma}
 % \email{minghuama@microsoft.com}
\affiliation{%
 \institution{Microsoft}
 \city{Beijing}
 \country{China}
 }

\author{Xin Wang}
 % \email{x.wang@stonybrook.edu}
\affiliation{%
 \institution{Stony Brook University}
 \city{New York}
 \country{USA}
 }

\author{Zhihan Li}
 % \email{lizhihan03@kuaishou.com}
\affiliation{%
 \institution{Kuaishou Technology}
 \city{Beijing}
 \country{China}
 }

\author{Dan Pei}
 % \email{peidan@tsinghua.edu.cn}
\affiliation{%
 \institution{Tsinghua University}
 \city{Beijing}
 \country{China}
 }

\author{Saravan Rajmohan}
% \email{saravan.rajmohan@microsoft.com}
\affiliation{%
 \institution{Microsoft}
 \city{Redmond}
 \country{USA}
 }

\author{Dongmei Zhang}
% \email{dongmeiz@microsoft.com}
\author{Qingwei Lin}
% \email{qlin@microsoft.com}
\affiliation{%
 \institution{Microsoft}
 \city{Beijing}
 \country{China}
 }

\author{Haiming Zhang}
% \email{hai@cnic.cn}
 \author{Jianhui Li}
   % \email{lijh@cnic.cn}
   \author{Gaogang Xie}
\affiliation{%
\institution{Computer Network Information Center, Chinese Academy of Sciences}
 \city{Beijing}
 \country{China}
 }

%   \author{Gaogang Xie}
%   % \email{xie@cnic.cn}
%   \affiliation{%
% \institution{Computer Network Information Center, Chinese Academy of Sciences}
%  \city{Beijing}
%  \country{China}
%  }

%  \author{Jianhui Li}
% \affiliation{%
% \institution{Computer Network Information Center, Chinese Academy of Sciences}
%  \city{Beijing}
%  \country{China}
%  }
%  \email{lijh@cnic.cn}

%  \author{Gaogang Xie}
% \affiliation{%
% \institution{Computer Network Information Center, Chinese Academy of Sciences}
%  \city{Beijing}
%  \country{China}
%  }
% \email{xie@cnic.cn}

\renewcommand{\shortauthors}{Wang et al.}
\begin{abstract}
Time series Anomaly Detection (AD) plays a crucial role for web systems. Various web systems rely on time series data to monitor and identify anomalies in real time, as well as to initiate diagnosis and remediation procedures. Variational Autoencoders (VAEs) have gained popularity in recent decades due to their superior de-noising capabilities, which are useful for anomaly detection. However, our study reveals that VAE-based methods face challenges in capturing long-periodic heterogeneous patterns and detailed short-periodic trends simultaneously. To address these challenges, we propose Frequency-enhanced Conditional Variational Autoencoder (FCVAE), a novel unsupervised AD method for univariate time series. To ensure an accurate AD, FCVAE exploits an innovative approach to concurrently integrate both the global and local frequency features into the condition of Conditional Variational Autoencoder (CVAE) to significantly increase the accuracy of reconstructing the normal data. Together with a carefully designed ``target attention'' mechanism, our approach allows the model to pick the most useful information from the frequency domain for better short-periodic trend construction. Our FCVAE has been evaluated on public datasets and a large-scale cloud system, and the results demonstrate that it outperforms state-of-the-art methods. This confirms the practical applicability of our approach in addressing the limitations of current VAE-based anomaly detection models.

%\hspace*{\fill}
%\noindent\textbf{Relevance Statement}: 

% \vspace{3mm}

% \noindent \textbf{Relevence Statement}: Our paper is highly relevant to the track ``Internet systems, applications, and Web of Things (WoT) applications.'' and previous papers in this conference\cite{www1,www2,www3,www4,www5,www6,www7,www8,donut}. It presents a novel method for anomaly detection in time series data, which is integral to the monitoring and real-time performance of various web and WoT systems, aligning with the track's focus on web performance, measurements, and characterization. Additionally, our work offers a new perspective on data management and stream processing for web applications, while also sharing experiences and lessons from the deployment of our innovative web-based algorithm.
\end{abstract}





\maketitle

\section{Introduction}

% \textbf{Relevance Statement: }Time series anomaly detection (AD) is ubiquitous for web systems \cite{www1,www2,www3,www4,www5,www6,www7,www8,donut}. Numerous web systems, such as online advertising systems, are monitored using a vast array of time series data (\textit{e.g.,} conversion rate) \cite{ctr}. Deploying time series AD algorithms is essential for timely detecting anomalies and initiating subsequent diagnosis and remediation processes. This paper primarily focuses on accurately detecting anomalies in web systems univariate time series (UTS).

Time series anomaly detection (AD) is ubiquitous for web systems \cite{www1,www2,www3,www4,www5,www6,www7,www8,donut,ganatra2023detection}. Numerous web systems, such as online advertising systems, are monitored using a vast array of time series data (\textit{e.g.,} conversion rate) \cite{ctr}. Deploying time series AD algorithms is essential for timely detecting anomalies and initiating subsequent diagnosis and remediation processes.

Anomalies are rare in real-world time series data \cite{rare}, making it difficult to label them and train a supervised model for anomaly detection \cite{donut}. Instead, unsupervised machine learning techniques are commonly used \cite{informer,ma2021jump,zhao2019automatic,zhao2023robust,dsanet,lstm,donut,begal,anotransfer,buzz,chen2023imdiffusion}. These techniques can be divided into two categories: \textit{prediction-based} \cite{informer,dsanet,lstm} and \textit{construction-based} \cite{donut,begal,anotransfer,buzz,ma2021jump}. Both types aim to identify normal values and compare them to actual values to detect anomalies. Prediction-based methods were originally developed for forecasting future data points, regardless of whether they were normal or anomalous. However, these methods may overfit to anomalous patterns and underperform. On the other hand, Variational AutoEncoders (VAEs) \cite{vae}, the leading construction-based approach, encode raw time series into a lower-dimensional latent space and then reconstruct them back to their original dimensions. VAEs are well-suited for detecting anomalies, but existing VAE-based anomaly detection models have not yet reached theoretically optimal performance. In this paper, we aim to re-examine the VAE model and improve its effectiveness in anomaly detection.

\begin{figure}[t]
\centerline{\includegraphics[scale=1]{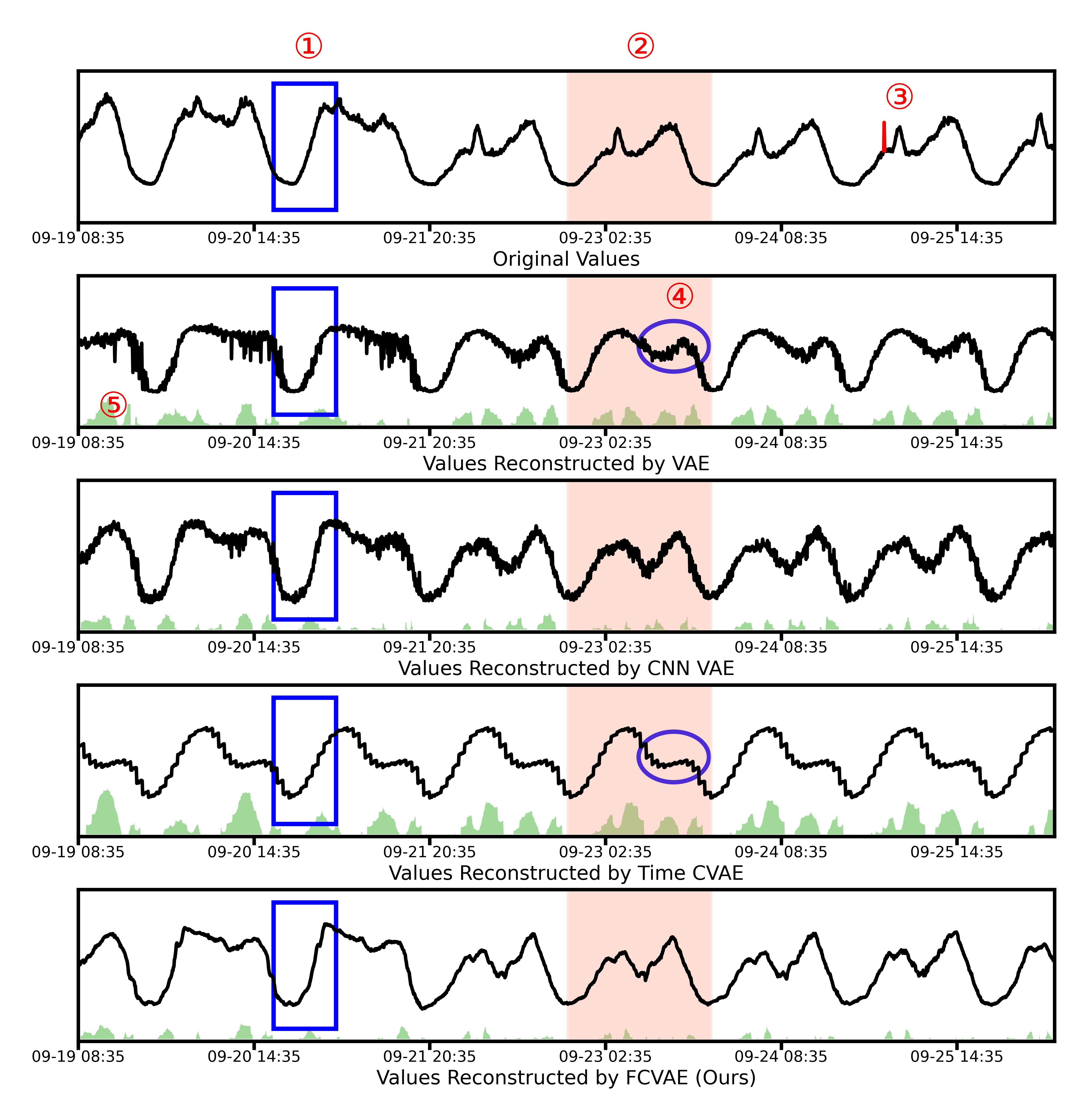}}
\caption{Comparison of four KPI reconstruction methods presented in our paper, highlighting anomalies in red \ding{174}. The green shade \ding{176} represents the difference between the reconstructed values and the original values, the red shade \ding{173} represents a long period, and the blue ellipse \ding{175} indicates peaks and valleys that are not properly reconstructed, the blue rectangle \ding{172} will be magnified in Figure~\ref{case} for detailed comparison.}
\label{overall-contrast}
\end{figure}

In order to more effectively demonstrate the challenges associated with VAE-based techniques, we provide an example in Figure~\ref{overall-contrast}. The original curve is displayed in the first sub-figure, with anomalies highlighted in red \ding{174}. The subsequent four sub-figures represent curves reconstructed by four distinct VAE-methods, including our proposed method (referred to as FCVAE). The reconstruction error is indicated by the green shaded area \ding{176}. To achieve superior AD performance, the reconstructed result should closely resemble the original curve for normal points, while deviating significantly for anomalous points \ding{174}. As evident in the figure, all VAE-based methods successfully disregard the anomalies during reconstruction. However, the reconstruction results for some normal points, particularly those marked by a blue rectangle \ding{172} and ellipse \ding{175}, are not satisfactory. This substantially impacts the overall performance, leading us to identify three key challenges that we address in the subsequent sections.

\begin{figure}[tbp]
\centering
\subfigure[Original Values\label{case:a}]{\includegraphics[scale=0.9]{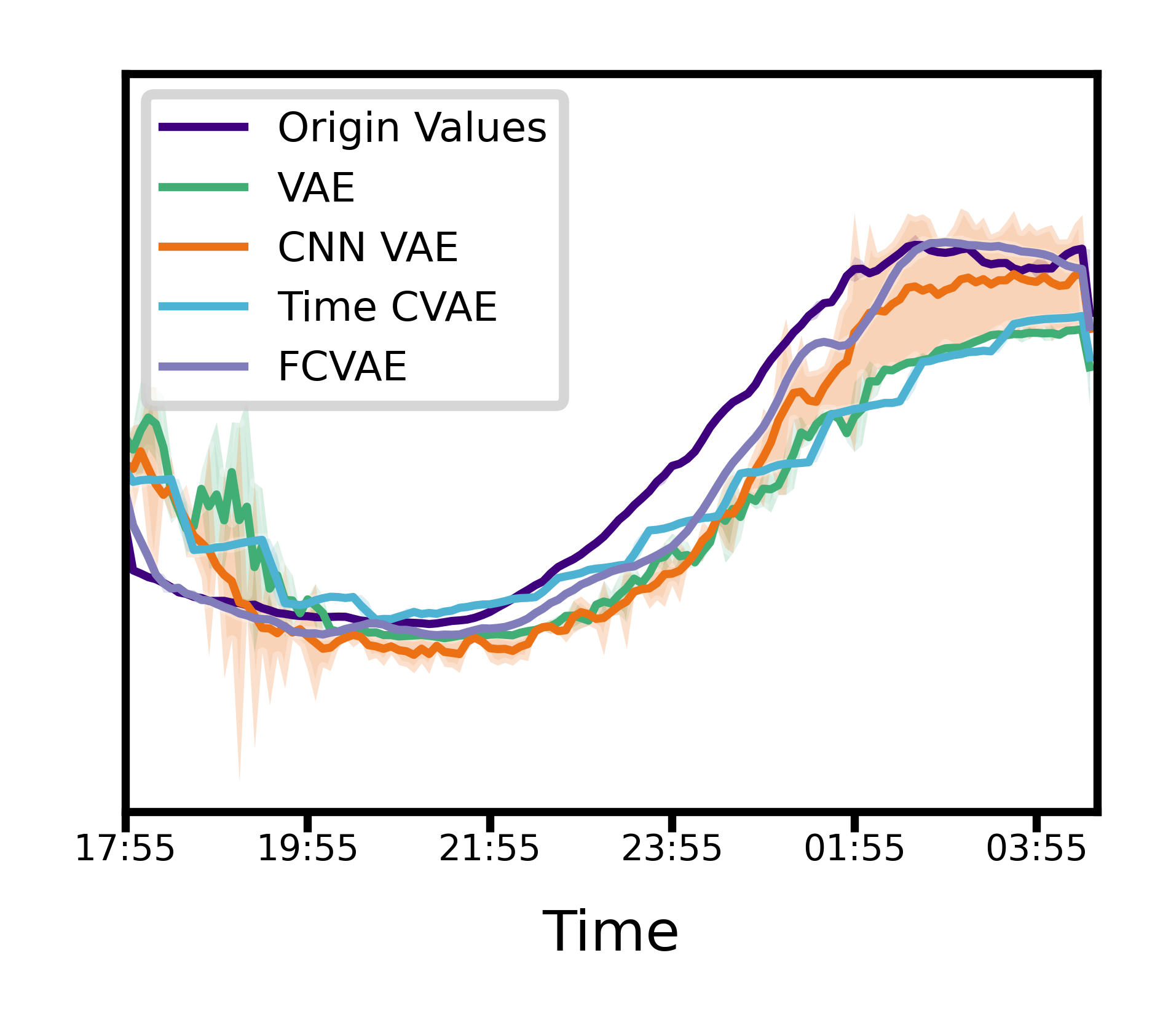}} \subfigure[Transformed Spectrogram\label{case:b}]{\includegraphics[scale=0.9]{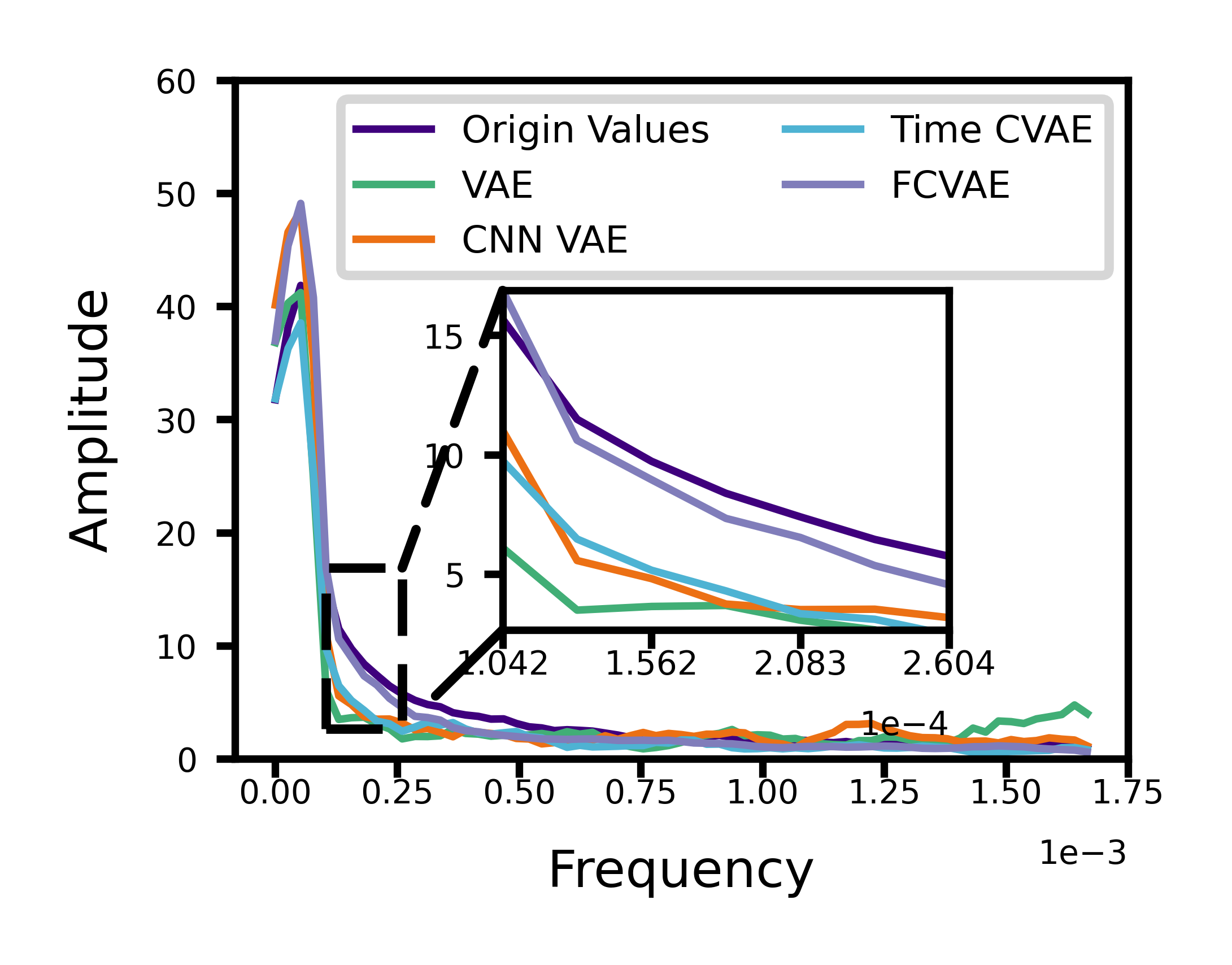}} 
\caption{A detailed view of the region enclosed by a blue rectangle \ding{172} in Figure~\ref{overall-contrast}, where the shaded area represents the value range before applying a sliding window average.}
\label{case}
\end{figure}

% vae's shortcomings
% cvae's shortcommings
% our method? 
\vspace{1mm}
\noindent\textbf{Challenge 1: Capturing similar yet heterogeneous periodic patterns.} From Figure~\ref{overall-contrast}, periodic patterns can be observed in the curves, with one such period emphasized by the red shaded area \ding{173}. However, the shapes across different periods vary. As demonstrated by the blue ellipse, existing VAE-based methods (as shown in the second sub-figure) are unable to capture these heterogeneous patterns effectively.
This observation naturally leads to the idea of utilizing conditional VAE to map data into distinct Gaussian spaces by considering the timestamp as a condition. Unfortunately, as illustrated in the fourth sub-figure (Time CVAE \cite{begal}), the results are unsatisfactory, which we will further discuss below.

\vspace{1mm}
\noindent\textbf{Challenge 2: Capturing detailed trends.} Reconstructing monotonous patterns (\textit{i.e.,} trends) might appear straightforward at first glance. However, upon a closer examination of the local window (highlighted in a blue rectangle \ding{172} in Figure~\ref{overall-contrast} and magnified in Figure~\ref{case:a}, it becomes evident that existing methods fail to restore detailed patterns within this time frame. In Figure~\ref{case:a}, the two green lines initially overestimate the ground truth (purple curve) but subsequently underestimate it for the remainder of the window. This is primarily because existing methods aim to minimize the overall reconstruction error without focusing on ``point-to-point'' dependencies, \textit{e.g.,} the precise upward and downward ranges following a specific point. This omission results in fluctuating reconstruction outcomes (as seen in the second sub-figure). Although CNN attempts to model point-to-point dependencies within the window, it still produces coarse-grained fluctuations (visible in the third sub-figure in Figure~\ref{overall-contrast}). The reason of unsatisfactory result of CNN-CVAE lies in Figure~\ref{case:b}. Upon converting the curves reconstructed by various methods (Figure~\ref{case:b}), it becomes evident that the primary cause of these phenomena is the \textbf{absence of some frequencies} (smaller amplitude of certain frequencies) in existing methods, hindering the reconstruction of detailed patterns. This observation logically suggests the possibility of employing frequency as the conditional factor in a Conditional Variational Autoencoder (CVAE). Nonetheless, employing frequency as the condition in CVAE presented a new challenge.

\vspace{1mm}
\noindent\textbf{Challenge 3: A large number of sub-frequencies make the signal in condition of CVAE noisy and difficult to use.} 
Directly converting the entire window into the frequency-domain results in numerous sub-frequencies, adding noise and obstructing effective VAE-based reconstruction. To address these challenges, we sub-divide the entire window into smaller ones and propose a \textbf{target attention} method to select the most useful sub-window frequencies.

In this paper, we introduce a novel unsupervised anomaly detection algorithm, named FCVAE (Frequency-enhanced Conditional Variational AutoEncoder). Different from current VAE-based anomaly detection methods, FCVAE innovatively incorporates both global and local frequency information to guide the encoding-decoding procedures, that both heterogeneous periodic and detailed trend patterns can be effectively captured. This in turn enables more accurate anomaly detection. Our paper's contributions can be summarized as follows:
\begin{itemize}%[leftmargin=*]
    \item Our analysis of the widely-used VAE model for anomaly detection reveals that existing VAE-based models fail to capture both heterogeneous periodic patterns and detailed trend patterns. We attribute this failure to the missing of some frequency-domain information, which current methods fail to reconstruct.
    \item Our study systematically improves the long-standing VAE by focusing on frequency. Our proposed FCVAE makes the VAE-based approach the state-of-the-art in anomaly detection once more. This is significant because VAE-based methods can inherently handle mixed anomaly-normal training data, while prediction-based methods cannot.
    \item Evaluations demonstrate that our FCVAE substantially surpasses state-of-the-art methods ($\sim$40\% on public datasets and 10\% in \textbf{a real-world web system} in terms of F1 score). Comprehensive ablation studies provide an in-depth analysis of the model, revealing the reasons behind its superior performance.
\end{itemize}

The replication package for this paper, including all our data,
source code, and documentation, is publicly available online at
\textbf{\url{https://github.com/CSTCloudOps/FCVAE}}.

\section{Preliminaries}
\subsection{Problem Statement}
Given a UTS $\mathbf{x} = [x_0, x_1, x_2, \cdots, x_t]$ and label series $\mathbf{L} = [l_0, l_1, l_2, \cdots, l_t]$, where $x_i \in \mathbb{R}$, $l_i \in \{0, 1\}$, and $t \in \mathbb{N}$. $\mathbf{x}$ represents the entire time series data array, while $x_i$ signifies the metric value at time $i$. $\mathbf{L}$ denotes the label of time series $\mathbf{x}$. We define the UTS anomaly detection task as follows: 

\emph{Given a UTS $\mathbf{x} = [x_0, x_1, x_2, \cdots, x_t]$, the objective of UTS anomaly detection is to utilize the data $[x_0, x_1, \cdots, x_{i-1}]$ preceding each point $x_i$ to predict $l_i$.} 
% Based on the value of $l_i$, we can determine whether $x_i$ is an anomaly or not.

% \subsection{Time domain and Frequency domain}
% \begin{figure}[htbp]
% \centerline{\includegraphics[scale=0.25]{pic/tfdomain.png}}
% \caption{Time domain and frequency domain.}
% \label{tfdomain}
% \end{figure}

% Time domain signals capture signal changes over time, providing specific values at each moment for practical applications. However, this representation alone lacks a comprehensive understanding of the signal, particularly regarding repeating patterns. Frequency domain information, illustrated in Figure~\ref{tfdomain}, offers an alternate perspective, resembling viewing an object from different angles. Combining frequency domain details unveils the spectrum, where a stable signal exhibits fewer components, concentrating energy, while a random signal presents more components, resulting in a broader spectrum. The Fast Fourier Transform (FFT) efficiently transforms signals from the time domain to the frequency domain.

\subsection{VAEs and CVAEs}
VAE is composed of an encoder $q_\phi(\mathbf{z|x})$ and a decoder $p_\theta(\mathbf{z|x})$. VAE can be trained by using the reparameterization trick. SGVB \cite{stochastic} is a commonly used training method for VAE because of its simplicity and effectiveness. It maximizes the evidence lower bound (ELBO) to simultaneously train the reconstruction and generation capabilities of VAE. 

DONUT \cite{donut} proposed the modified ELBO (M-ELBO) to weaken the impact of abnormal and missing data in the window on the reconstruction. M-ELBO is defined in \eqref{melbo}, $\alpha_w$ is defined as an indicator, where $\alpha_w = 1$ indicates $x_w$ being not anomalous or missing, and $\alpha_w = 0$ otherwise. $\beta$ is defined as $(\sum_{w=1}^{W}\alpha_w) / W$.

% The ELBO is defined in \eqref{elbo}.

% \begin{footnotesize}
% \begin{equation}
% \begin{aligned}
% &\mathcal{L}= \mathbb{E}_{q_\phi(\mathbf{z|x})}[\mathrm{log} p_\theta(\mathbf{x|z})+\mathrm{log} p_\theta(\mathbf{z})-\mathrm{log} q_\phi(\mathbf{z|x})]
% \label{elbo}
% \end{aligned}
% \end{equation}
% \end{footnotesize}

\begin{footnotesize}
\begin{equation}
\begin{aligned}
\mathcal{L} = \mathbb{E}_{q_\phi(\mathbf{z|x})}[\sum_{w=1}^{W}\alpha_{w}\mathrm{log} p_\theta(x_w|\mathbf{z})+\beta\mathrm{log} p_\theta(\mathbf{z|x})-\mathrm{log} q_\phi(\mathbf{z|x})]
\label{melbo}
\end{aligned}
\end{equation}
\end{footnotesize}

The overall structure of CVAE \cite{cvae} is similar to VAE, and it combines conditional generative models with VAE to achieve stronger control over the generated data. 
% CVAE introduces the concept of condition, which plays a role analogous to labels within the model. The encoder part of CVAE attempts to learn $q_\phi(\mathbf{z|x,c})$, meaning that the input data $\mathbf{x}$ and its condition $\mathbf{c}$ are fed into the encoder together. The decoder part of CVAE aims to learn $p_\theta(\mathbf{x|z,c})$, and both the latent variable $\mathbf{z}$ and the condition are combined when passing through it. 
The training objective of CVAE is defined as \eqref{closs}, where $\mathbf{c}$ is the condition, similar to that of VAE. FCVAE which will be elaborated on later extends the CVAE framework by incorporating frequency information.

\begin{footnotesize}
\begin{equation}
\begin{aligned}
 \mathcal{L}&= \mathbb{E}_{q_\phi(\mathbf{z|x,c})}[\mathrm{log} p_\theta(\mathbf{x|z,c})+\mathrm{log} p_\theta(\mathbf{z})-\mathrm{log} q_\phi(\mathbf{z|x,c})]
\label{closs}
\end{aligned}
\end{equation}
\end{footnotesize}

\section{Methodology}
% In this section, we first introduce the overall framework and FCVAE model structure that can be used to improve anomaly detection performance. Next, we will discuss the overall training and testing process.

\subsection{Framework Overview}

\begin{figure}[htbp]
\centerline{\includegraphics[scale=0.30]{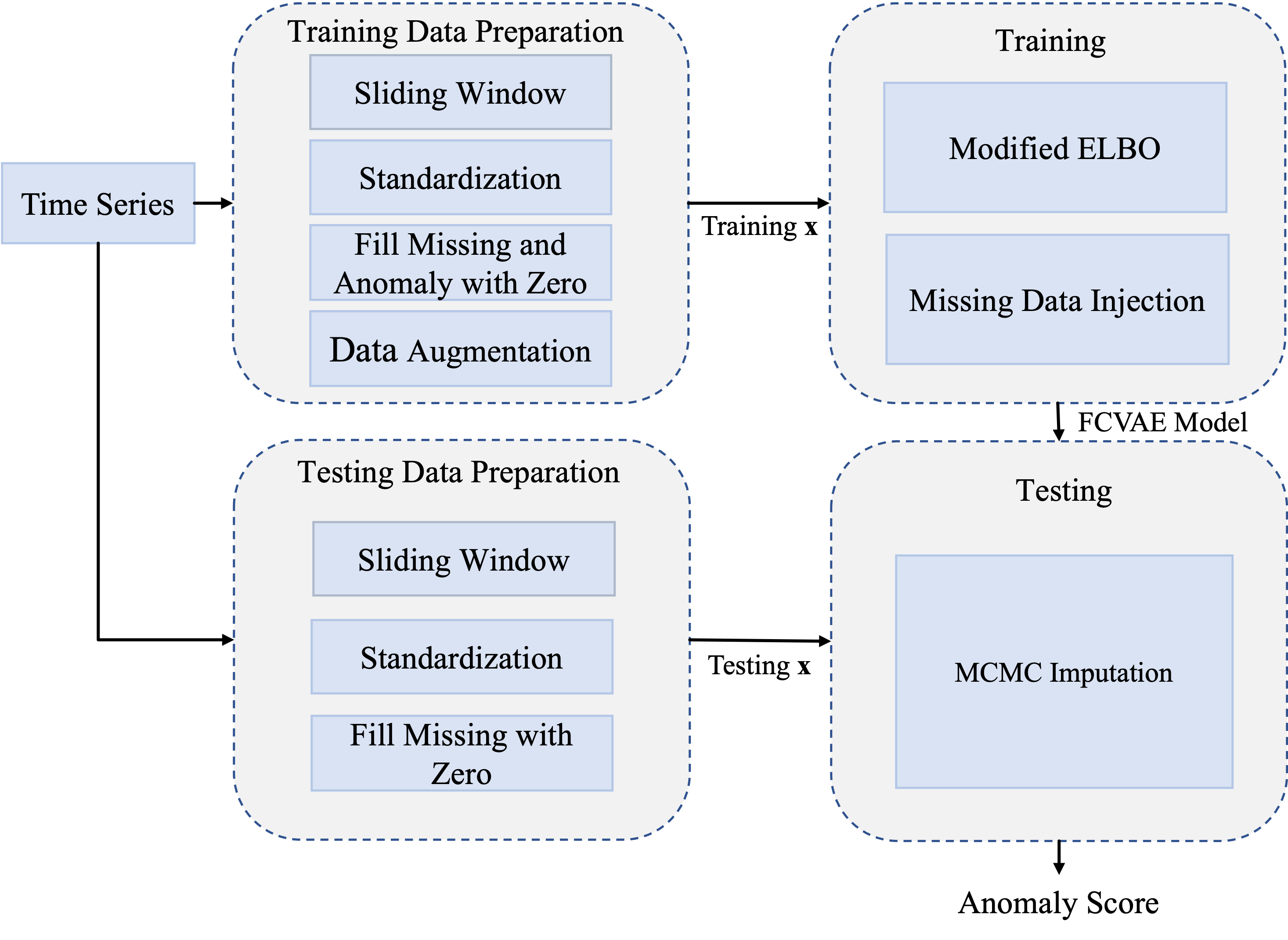}}
\caption{Overall Framework.}
\label{framework}
\end{figure}

The proposed algorithm for anomaly detection is illustrated in Figure~\ref{framework} and comprises three main components: data preprocessing, training, and testing.

% Given a data point $x_t$, as defined in the problem statement, its state can only be evaluated based on its preceding points $[x_0,x_1,\cdots,x_t]$. To maintain model consistency, a sliding window method is employed, where a window of $W$ consecutive points, $[x_{t-W+1},x_{t-W+2},\cdots,x_t]$, is utilized to determine if $x_t$ is anomalous. Following sliding window and data preprocessing, a batch of data is input into the FCVAE model for offline training, which will be presented in detail later. Subsequently, the trained model is transferred to the online test module for testing and computing the anomaly score.

\begin{figure*}[t]
\centerline{\includegraphics[scale=0.5]{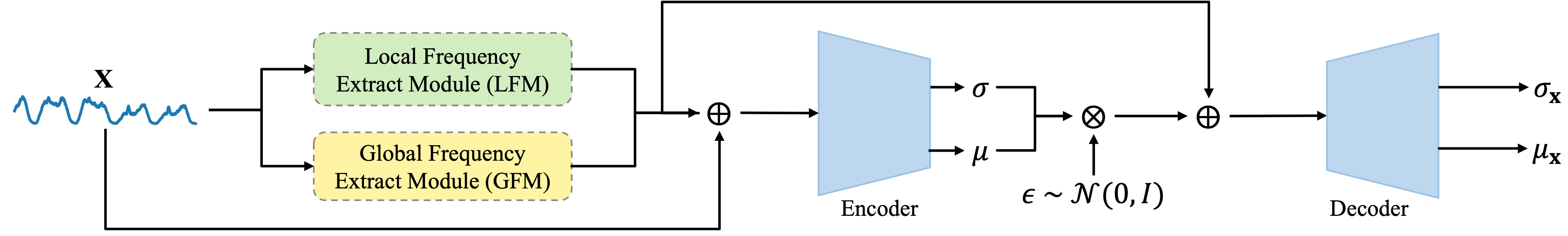}}
\caption{FCVAE Model Architecture.}
\label{model}
\end{figure*}

\begin{figure*}[htbp]
\centerline{\includegraphics[scale=0.45]{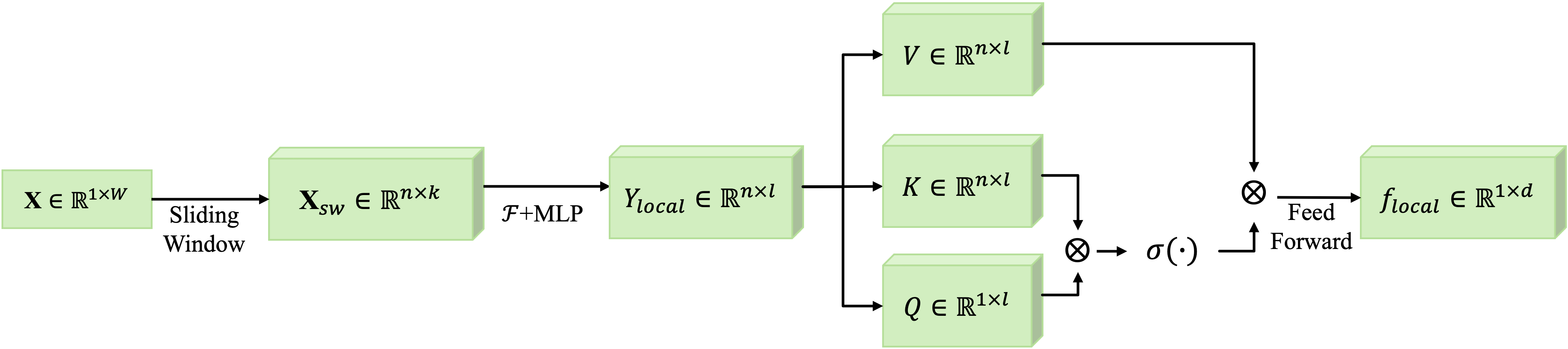}}
\caption{Architecure of LFM.}
\label{LFM}
\end{figure*}

\subsection{Data Preprocessing}
Data preprocessing encompasses standardization, filling missing and anomaly points, and the newly introduced method of \textbf{data augmentation}. The efficacy of data standardization and filling missing and anomaly points has been substantiated in prior studies \cite{donut, interfusion, begal}. Therefore, we directly incorporate these techniques into our approach.

\begin{figure}[htbp]
    \centerline{
    \subfigure[Pattern Anomaly\label{anomaly:a}]{\includegraphics[scale=0.7]{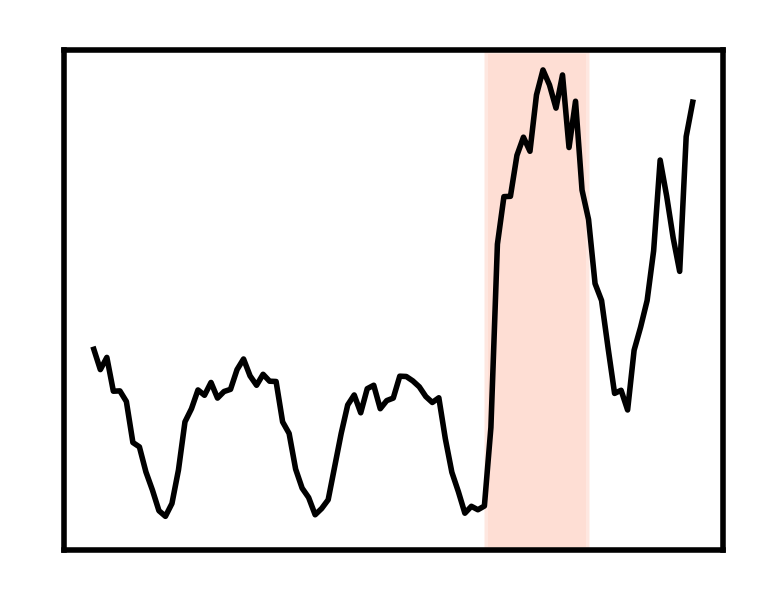}}
    \subfigure[Value Anomaly\label{anomaly:b}]{\includegraphics[scale=0.7]{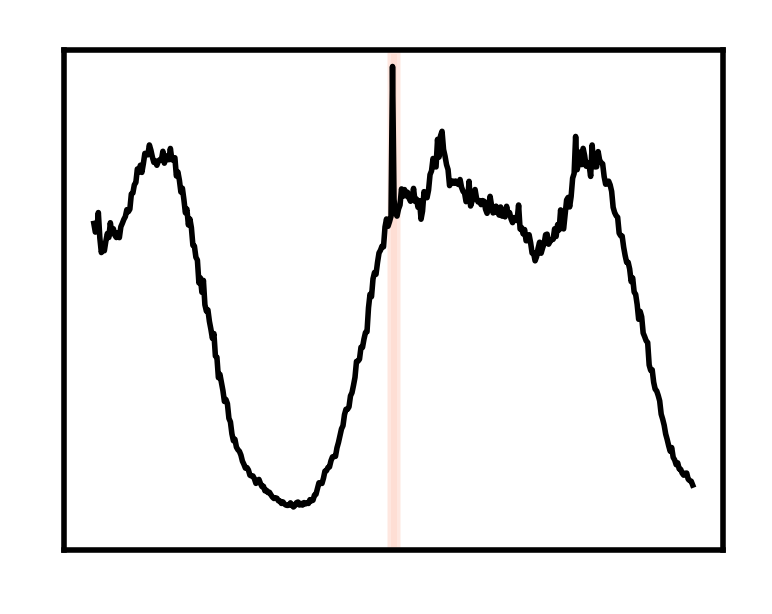}}
    }
    \caption{Examples of the two most frequent anomalies, where the red shaded area denotes the abnormal segments.}
    \label{anomaly}
\end{figure}

Previous data augmentation methods \cite{tfad,dataaug1,dataaug2} often added normal samples, such as variations of data from the time domain or frequency domain. However, for our method, we train the model by incorporating all the time series from the dataset together, which provides sufficient pattern diversity. Furthermore, FCVAE has the ability to extract pattern information due to the addition of frequency information, so it can handle new patterns well. Nonetheless, even with the introduction of frequency information, anomalies are often challenging to be effectively addressed. For the model to learn how to handle anomalies, we primarily focus on abnormal data augmentation. In time series data, anomalies are mostly manifested as pattern mutations or value mutations (shown in Figure~\ref{anomaly}), so our data augmentation mainly targets to these two aspects. The augmentation on the pattern mutation is generated  by combining two windows from different curves, with the junction acting as the anomaly. Value mutation refers to changing some points in the window to randomly assigned abnormal values. With the augmented anomaly data, M-ELBO in CVAE, which will be introduced in detail later, can perform well even in an unsupervised setting without true labels.

\subsection{Network Architecture}
% Time series data often contain valuable frequency domain information, offering a distinct perspective in comparison to the time domain information. In the frequency domain, data can be represented as a combination of some periodic waveforms or frequency components, revealing information such as periodicity, seasonal changes, periodic trends, and cyclic noise in the data. Moreover, frequency domain information is crucial in several applications, including finance, where it can be employed to predict stock market volatility, currency exchange rate fluctuations, and economic trends. In signal and image processing, frequency domain information proves useful for tasks such as denoising, image enhancement, and feature extraction. Consequently, our model leverages the frequency information to more effectively reconstruct the time series data.

The proposed FCVAE model is illustrated in Figure~\ref{model}. It comprises three main components: encoder, decoder, and a condition extraction block that includes a global frequency information extraction module (GFM) and a local frequency information extraction module (LFM). Equation \eqref{eq5} illustrates how our model works.

% utilizes the LFM and GFM to extract both local and global frequency features from the input signal $\mathbf{x}$. These features, in conjunction with $\mathbf{x}$, are then fed into the encoder $p_\theta$ to derive the latent variable $\mathbf{z}$. The decoder $q_\phi$ subsequently maps $\mathbf{z}$, along with the earlier obtained frequency information, back to the data space, facilitating the reconstruction of the distribution of the input signal. The architecture of the Encoder-Decoder model is depicted in Figure~\ref{encoder}. This methodology effectively integrates frequency information into the reconstruction process, thereby augmenting the model's capacity to discern intricate patterns and anomalies within time series data.

\begin{footnotesize}
\begin{equation}
\begin{aligned}
&\mu,\sigma=\mathrm{Encoder}(\mathbf{x},\mathrm{LFM(\mathbf{x})},\mathrm{GFM(\mathbf{x})})\\
& \mathbf{z}=\mathrm{Sample} (\mu,\sigma)\\
&\mu_{\mathbf{x}},\sigma_{\mathbf{x}}=\mathrm{Decoder}(\mathbf{z},\mathrm{LFM(\mathbf{x})},\mathrm{GFM(\mathbf{x})})
\label{eq5}
\end{aligned}
\end{equation}
\end{footnotesize}

\subsubsection{GFM}

\begin{figure}[h]
\centerline{\includegraphics[scale=0.45]{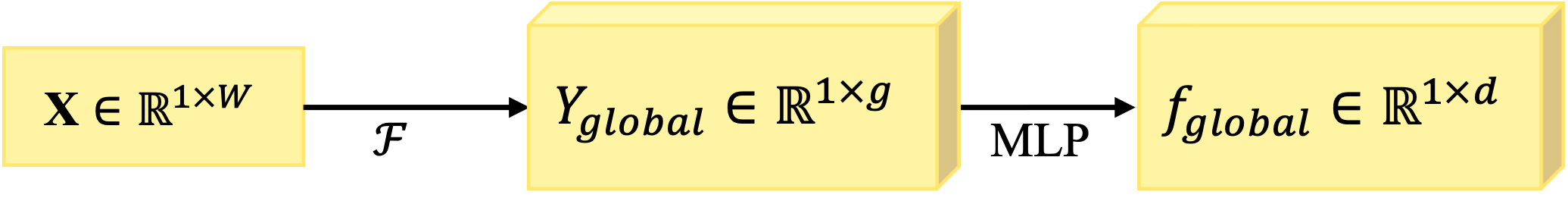}}
\caption{Architecure of GFM.}
\label{GFM}
\end{figure}

The GFM module (Figure~\ref{GFM}) extracts the global frequency information using the FFT transformation ($\mathcal{F}$). However, not all frequency information is useful. The frequencies resulted from the noise and anomalies in the time series data appear as long tails in the frequency domain. Therefore, we employ a linear layer after the FFT to filter out the useful frequency information that can represent the current window pattern. Moreover, we incorporate a dropout layer following Fedformer \cite{fedformer} to enhance the model's ability to learn the missing frequency information.

The $f_{global}\in\mathbb{R}^{1\times d}$ is calculated as \eqref{eq6}, where d is the embedding dimension of the global frequency information and $\mathcal{F}$ means FFT. 

\begin{footnotesize}
\begin{equation}
\begin{aligned}
f_{global} = \mathrm{Dropout}(\mathrm{Dense}(\mathcal{F}(\mathbf{x})))
\label{eq6}
\end{aligned}
\end{equation}
\end{footnotesize}

\subsubsection{LFM}
The attention mechanism \cite{attention} has been widely adopted in time series data processing due to its ability to dynamically process dependencies between different time steps and focus on important ones. Target attention, which is developed based on attention, is widely used in the field of recommendation \cite{targetattention}. Specifically, target attention can weigh the features of the target domain, leading to more accurate domain adaptation.

% \begin{footnotesize}
% \begin{equation}
% \begin{aligned}
% \mathrm{Atten}(\mathbf{q,k,v}) = \mathrm{Softmax}(\frac{\mathbf{qk}^\top}{\sqrt{d_q}})\mathbf{v}
% \label{eq8}
% \end{aligned}
% \end{equation}
% \end{footnotesize}

The GFM module extracts the frequency information from the entire window, proving to be effective in reconstructing the data within the whole window. However, we use a window to detect whether the last point is abnormal, which poses a challenge because the GFM module does not provide sufficient attention to the last point. This can result in a situation where the reconstruction is satisfactory for part of the window but not for another part, especially when changes in system services lead to the concept drift in the time series data. Even in the absence of concept drift, GFM cannot capture local changes as it extracts the average frequency information from the entire window; hence, the reconstruction of the last key point may be unsatisfactory. Nonetheless, as previously mentioned, target attention can effectively address this issue, as it captures the frequency information of the entire window while paying a greater attention to the latest time point. Therefore, we propose the LFM that incorporates the target attention.

As depicted in Figure~\ref{LFM}, the LFM module operates by sliding the entire window $\mathbf{x}$ to obtain several small windows $\mathbf{x}_{sw}$. Subsequently, FFT and frequency information extraction are applied to each small window. The most recent small window is used as the query $Q$ because it contains the last point that we want to detect. The remaining small windows are utilized as keys $K$ and values $V$ for target attention. Finally, a linear layer is employed to facilitate the model in learning to extract the most important and useful part of the local frequency information, and dropout is also applied to enhance the model's ability to reconstruct some of the local frequency information like GFM.

\begin{footnotesize}
\begin{equation}
\begin{aligned}
&\mathbf{x}_{sw} = \mathrm{SlidingWindow}(\mathbf{x})\\
&Q=\mathrm{Select}(\mathrm{Dense}(\mathcal{F}(\mathbf{x}_{sw})))\\
&K,V=\mathrm{Dense}(\mathcal{F}(\mathbf{x}_{sw}))\\
%,V=\mathrm{Dense}(\mathcal{F}(\mathbf{x}_{sw}))\\
&f_{local} = \mathrm{Dropout}(\mathrm{FeedFawrd}((\sigma(Q\cdot K^\top)\cdot V))
\label{eq7}
\end{aligned}
\end{equation}
\end{footnotesize}

The calculation of $f_{local} \in\mathbb{R}^{1\times d}$ in LFM is given by \eqref{eq7}, where $d$ is the embedding dimension of the local frequency information, which is the same as that of GFM. Here, $\mathbf{x}_{sw} \in \mathbb{R}^{n\times k}$ represents a group of small windows extracted from the original window, where $k$ is the dimension of the small windows and $n$ is the number of small windows. The Select function is employed to select the latest window as the query $Q$ and the Dense function means dense neural network. The softmax function $\sigma$ is used to calculate the attention weights for the small windows.

% \subsection{Encoder and Decoder}
% \begin{figure}[htbp]
% \centerline{\includegraphics[scale=0.4]{pic/Enco and deco.png}}
% \caption{Architecture of Encoder and Decoder.}
% \label{encoder}
% \end{figure}

% The architecture of the Encoder-Decoder model is depicted in Figure~\ref{encoder}. The computations performed by the encoder and decoder can be expressed as shown in \eqref{encoder_eq}. Here, $\phi$ represents the network parameters of the hidden layers in the encoder, while $\theta$ represents the network parameters of the hidden layers in the decoder. The \textbf{W} and \textbf{b} variables denote the parameters of the dense layer, which are also shown in Figure~\ref{encoder}. The SoftPlus function is defined as $\text{SoftPlus}(a) = \log(\exp(a)+1)$.

% \begin{footnotesize}
% \begin{equation}
% \begin{aligned}
% &\mathbf{c} = \mathrm{Concat}(\mathrm{LFM(\mathbf{x}),LFM(\mathbf{x})})\\
% &\mu=\mathbf{W}_{\mu}^{\top}f_{\phi}(\mathbf{x,c})+\mathbf{b}_{\mu}\\
% &\sigma=\mathrm{SoftPlus}[\mathbf{W}_{\sigma}^{\top}f_{\phi}(\mathbf{x,c})+\mathbf{b}_{\sigma}]\\
% &\mu_{\mathbf{x}}=\mathbf{W}_{\mu_{\mathbf{x}}}^{\top}f_{\theta}(\mathbf{z,c})+\mathbf{b}_{\mu_{\mathbf{x}}}\\
% &\sigma_{\mathbf{x}}=\mathrm{SoftPlus}[\mathbf{W}_{\sigma_{\mathbf{x}}}^{\top}f_{\theta}(\mathbf{z,c})+\mathbf{b}_{\sigma_{\mathbf{x}}}]
% \label{encoder_eq}
% \end{aligned}
% \end{equation}
% \end{footnotesize}

\subsection{Training and Testing}
The training process of FCVAE incorporates three key technologies: CVAE-based modified evidence lower bound (CM-ELBO), missing data injection, as well as the newly proposed \textbf{masking the last point}. As shown in \eqref{cmelbo}, \textbf{CM-ELBO} is obtained by applying M-ELBO \cite{donut} to CVAE. Missing data injection \cite{donut,interfusion} is a commonly used technique in VAE that we directly apply. We observed that an anomalous point in time series data manifests as an outlier value in the time domain. However, when the data are transformed into the frequency domain, all frequency information is shifted, leading to a challenge. The impact of this issue will be amplified when the last point is abnormal, as we specifically aim to detect the last point given the whole window. While we use the frequency enhancement method and frequency selection to mitigate this problem to some extent, we mask the last point as zero during the extraction of the frequency condition to address this issue further.

\begin{footnotesize}
\begin{equation}
\begin{aligned}
\mathcal{L} = \mathbb{E}_{q_\phi(\mathbf{z|x,c})}[\sum_{w=1}^{W}\alpha_{w}\mathrm{log} p_\theta(x_w|\mathbf{z,c})+\beta\mathrm{log} p_\theta(\mathbf{z})-\mathrm{log} q_\phi(\mathbf{z|x,c})]
\label{cmelbo}
\end{aligned}
\end{equation}
\end{footnotesize}

While testing, FCVAE adopts the Markov Chain Monte Carlo (MCMC)-based missing imputation algorithm proposed in \cite{stochastic} and applied in \cite{interfusion} to mitigate the impact of missing data. Since our goal is to detect the last point of a window, the last point is set to missing for MCMC to obtain a normal value. This also allows for a better adaptation to the last point mask mentioned earlier. FCVAE further utilizes reconstruction probabilities as anomaly scores, which are defined in equation \eqref{recon}.

\begin{footnotesize}
\begin{equation}
\begin{aligned}
\mathrm{Anomaly Score}=-\mathbb{E}_{q_\phi\mathbf{(z|x,c)}}[\mathrm{log}p_\theta(\mathbf{x|z,c})]
\label{recon}
\end{aligned}
\end{equation}
\end{footnotesize}

\section{Experiments}
% In this section, we first introduce the experiment settings. Then, we conduct experiments to evaluate our proposed FCVAE method based on the following six research questions.

% \begin{itemize}%[leftmargin=*]
%     \item \textbf{RQ1:} How does our FCVAE method perform in comparison to state-of-the-art methods on commonly used public datasets?
%     \item \textbf{RQ2:} In the field of UTS anomaly detection, various types of information can be encoded as conditions for CVAE. Which type of information is most effective as a condition for CVAE?
%     \item \textbf{RQ3:} Frequency information is crucial for time series data. How can a model structure be designed to better and more comprehensively utilize frequency information?
%     \item \textbf{RQ4:} Is employing global and local frequency information as conditions for CVAE truly beneficial for reconstructing normal data and detecting anomalies?
%     \item \textbf{RQ5:} Does the attention mechanism in LFM function effectively?
%     \item \textbf{RQ6:} Are the key techniques mentioned in the architecture advantageous for anomaly detection?
% \end{itemize}

\subsection{Experiment Settings}
\subsubsection{Datasets}
To evaluate the effectiveness of our proposed algorithm, we conducted experiments on four datasets. \textbf{Yahoo} \cite{yahoo} is an open data set for anomaly detection released by Yahoo lab. \textbf{KPI} \cite{aiops} KPI is collected from five large Internet companies (Sougo, eBay, Baidu, Tencent, and Ali). \textbf{WSD} \cite{wsd2} Web service dataset (WSD) contains real-world KPIs collected from three top-tier Internet companies, Baidu, Sogou, and eBay, providing large-scale Web services. \textbf{NAB} \cite{nab} The Numenta Anomaly Benchmark (NAB) is an open dataset created by Numenta company for evaluating the performance of time series anomaly detection algorithms.
% \begin{table}[htbp]
% \caption{Statistics of datasets}
% \begin{center}
% \begin{tabular}{cccc}
% \toprule
% DataSet & Total Curves &Total Points & Anomaly Points \\ \midrule
% Yahoo & 367 & 572966 &  3896/0.68\% \\
% KPI & 58 & 5922913 &  134114/2.26\% \\
% WSD & 210 & 7592905 & 121279/1.59\%\\
% NAB & 10 &158621 & 15684/9.88\% \\
% \bottomrule
% \end{tabular}
% \label{dataset}
% \end{center}
% \end{table}

\subsubsection{Baseline Methods}
To benchmark our model FCVAE against existing methods, we chose the following approaches for evaluation: SPOT \cite{spot}, SRCNN \cite{srcnn}, TFAD \cite{tfad}, DONUT \cite{donut}, Informer \cite{informer}, Anomaly-Transformer \cite{anomaly-transformer}, AnoTransfer \cite{anotransfer}, VQRAE \cite{vqrae}. SPOT represents a traditional statistical method rooted in extreme value theory. SRCNN and TFAD are supervised methods relying on high-quality labels. Donut, VQRAE, and AnoTransfer are unsupervised reconstruction-based methods utilizing VAE for normal value reconstruction. Informer is an unsupervised prediction-based method that endeavors to predict normal values using an attention mechanism. Anomaly-Transformer is an unsupervised anomaly detection method leveraging the transformer architecture. 

\subsubsection{Evaluation Metrics}
In practical applications, operators tend to be less concerned with point-wise anomaly detection, i.e., whether each individual point is classified as anomalous or not, and focus more on detecting continuous anomalous segments in time series data. Moreover, due to the substantial impact of anomalous segments, operators aim to identify such segments as early as possible. To address these requirements, we adopt two metrics, best F1 and delay F1, which are based on the works of DONUT \cite{donut} and SRCNN \cite{srcnn}, respectively.

\begin{figure}[htbp]
\centerline{\includegraphics[scale=0.35]{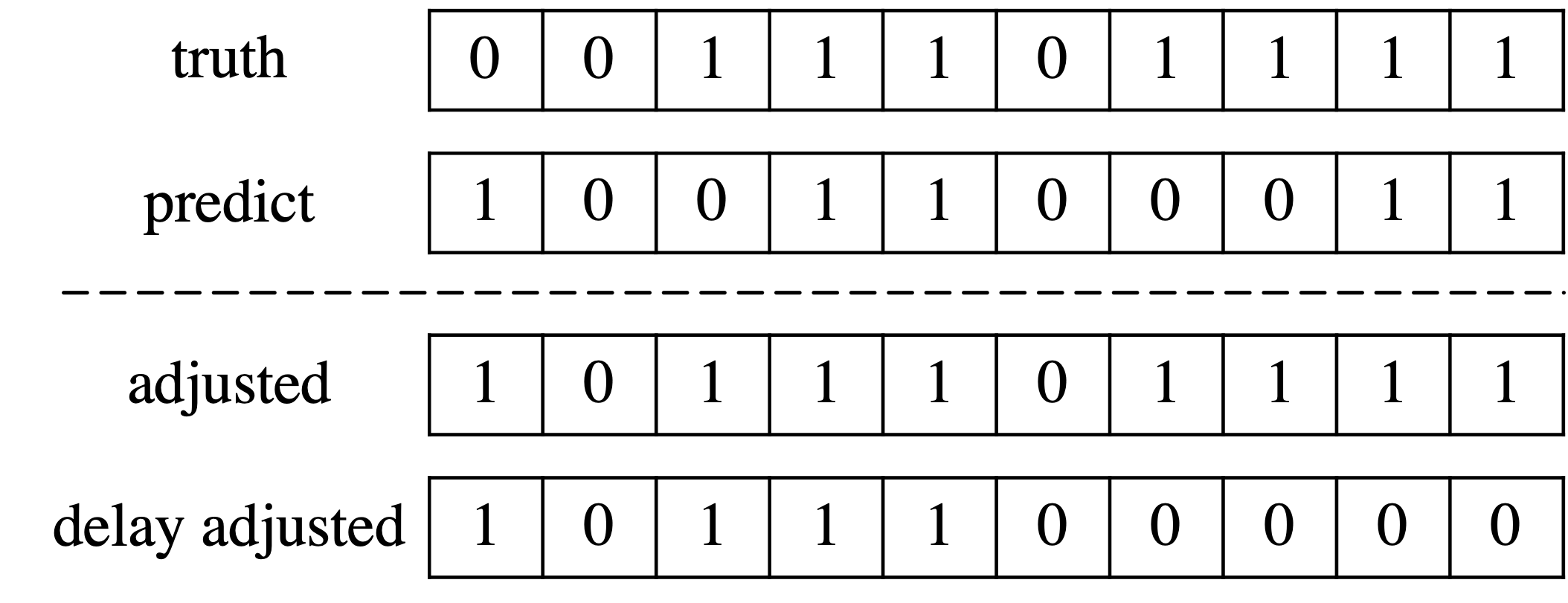}}
\caption{Illustration of the adjustment strategy.}
\label{pointadjust}
\end{figure}

% \begin{table*}[htbp]
% \footnotesize
% \caption{Performance on test data.}
% \begin{center}
% \resizebox{\textwidth}{!}{
% \begin{tabular}{c|cc|cc|cc|cc}
% \toprule
%      & \multicolumn{2}{|c|}{\textbf{Yahoo}} & \multicolumn{2}{|c|}{\textbf{KPI}} & \multicolumn{2}{|c|}{\textbf{WSD}} & \multicolumn{2}{|c}{\textbf{NAB}} \\

% \textbf{Method} &   \textbf{Best F1}&\textbf{Delay F1}  & \textbf{Best F1}&\textbf{Delay F1}  & \textbf{Best F1}&\textbf{Delay F1}  & \textbf{Best F1}&\textbf{Delay F1} \\ \midrule

% \textbf{SPOT} \cite{spot} &0.417&0.417& 0.360& 0.143&0.472 &0.237 &0.829&0.829 \\

% \textbf{SRCNN} \cite{srcnn} &0.251&0.198&0.786&0.678 &0.170 &0.053  &0.828&0.575 \\

% \textbf{DONUT} \cite{donut} &0.215&0.215 &0.454 &0.364  & 0.224&0.158 &0.935&0.797 \\
% \textbf{VQRAE} \cite{vqrae}&0.510&0.492&0.272&0.137& 0.312&0.103 &0.933&0.893 \\

% \textbf{Anotransfer} \cite{anotransfer} &0.567&0.496&0.685&0.461&0.674&0.379&0.965&0.871  \\
% \textbf{Informer} \cite{informer}  &0.707&0.671&\textbf{\underline{0.918}}&\textbf{\underline{0.822}}&0.557&0.393&\textbf{\underline{0.973}}&0.892 \\
% \textbf{TFAD} \cite{tfad} &\textbf{\underline{0.805}}&\textbf{\underline{0.802}}&0.752&0.680&0.628&\textbf{\underline{0.455}}&0.734&0.248\\

% \textbf{Anomaly-Transformer} \cite{anomaly-transformer} &0.274&0.029&0.868&0.346&\textbf{\underline{0.728}}&0.137&0.971&\textbf{\underline{0.911}}\\
% \textbf{FCVAE}  &\textbf{0.857}&\textbf{0.842}&\textbf{0.927}&\textbf{0.835}&\textbf{0.831}&\textbf{0.631}&\textbf{0.976}&\textbf{0.917}\\
% \bottomrule
% \end{tabular}
% }
% \label{bestf1}
% \end{center}
% \end{table*}

\begin{table*}
\centering
\caption{Performance on test data. P means precison, R means recall, F1 means best F1 and F1* means delay F1.}
\label{bestf1}
\setlength{\tabcolsep}{2.2pt}
\resizebox{\linewidth}{!}{%
\begin{tabular}{c|cccccc|cccccc|cccccc|cccccc} 
\toprule
                             & \multicolumn{6}{c|}{\textbf{Yahoo}}                                                                 & \multicolumn{6}{c|}{\textbf{KPI}}                                                                   & \multicolumn{6}{c|}{\textbf{WSD}}                                                                   & \multicolumn{6}{c}{\textbf{NAB}}                                                                     \\
\textbf{Method}              & \textbf{P} & \textbf{R} & \textbf{F1}            & \textbf{P} & \textbf{R} & \textbf{F1*}           & \textbf{P} & \textbf{R} & \textbf{F1}            & \textbf{P} & \textbf{R} & \textbf{\textbf{F1*}}  & \textbf{P} & \textbf{R} & \textbf{F1}            & \textbf{P} & \textbf{R} & \textbf{\textbf{F1*}}  & \textbf{P} & \textbf{R} & \textbf{F1}            & \textbf{P} & \textbf{R} & \textbf{\textbf{F1*}}   \\ 
\midrule
\textbf{SPOT\cite{spot}}                & 0.572      & 0.328      & 0.417                  & 0.572      & 0.328      & 0.417                  & 0.966      & 0.221      & 0.360                  & 0.911      & 0.077      & 0.143                  & 0.947      & 0.315      & 0.472                  & 0.887      & 0.137      & 0.237                  & 0.992      & 0.713      & 0.829                  & 0.992      & 0.713      & 0.829                   \\
\textbf{SRCNN\cite{srcnn}}               & 0.268      & 0.236      & 0.251                  & 0.219      & 0.181      & 0.198                  & 0.673      & 0.944      & 0.786                  & 0.617      & 0.753      & 0.678                  & 0.093      & 0.903      & 0.170                  & 0.028      & 0.361      & 0.053                  & 0.825      & 0.832      & 0.828                  & 0.460      & 0.766      & 0.575                   \\
\textbf{DONUT\cite{donut}}               & 0.381      & 0.150      & 0.215                  & 0.381      & 0.150      & 0.215                  & 0.378      & 0.569      & 0.454                  & 0.328      & 0.407      & 0.364                  & 0.263      & 0.195      & 0.224                  & 0.199      & 0.131      & 0.158                  & 0.933      & 0.937      & 0.935                  & 0.821      & 0.774      & 0.797                   \\
\textbf{VQRAE\cite{vqrae}}               & 0.706      & 0.399      & 0.510                  & 0.691      & 0.381      & 0.492                  & 0.202      & 0.418      & 0.272                  & 0.167      & 0.117      & 0.137                  & 0.518      & 0.233      & 0.312                  & 0.241      & 0.066      & 0.103                  & 0.990      & 0.882      & 0.933                  & 0.806      & 1.000      & 0.893                   \\
\textbf{Anotransfer\cite{anotransfer}}         & 0.902      & 0.413      & 0.567                  & 0.575      & 0.437      & 0.496                  & 0.815      & 0.591      & 0.685                  & 0.557      & 0.394      & 0.461                  & 0.695      & 0.654      & 0.674                  & 0.331      & 0.444      & 0.379                  & 0.962      & 0.968      & 0.965                  & 0.837      & 0.908      & 0.871                   \\
\textbf{Informer\cite{informer}}            & 0.747      & 0.671      & 0.707                  & 0.731      & 0.619      & 0.671                  & 0.927      & 0.910      & \textbf{\underline{0.918}} & 0.801      & 0.845      & \textbf{\underline{0.822}} & 0.532      & 0.583      & 0.557                  & 0.402      & 0.385      & 0.393                  & 0.971      & 0.974      & \textbf{\underline{0.973}} & 0.878      & 0.907      & 0.892                   \\
\textbf{TFAD\cite{tfad}}                & 0.884      & 0.739      & \textbf{\underline{0.805}} & 0.883      & 0.734      & \textbf{\underline{0.802}} & 0.684      & 0.834      & 0.752                  & 0.650      & 0.714      & 0.680                  & 0.541      & 0.750      & 0.628                  & 0.431      & 0.482      & \textbf{\underline{0.455}} & 0.749      & 0.719      & 0.734                  & 0.265      & 0.233      & 0.248                   \\
\textbf{Anomaly-Transformer\cite{anomaly-transformer}} & 0.588      & 0.179      & 0.274                  & 0.054      & 0.020      & 0.029                  & 0.930      & 0.814      & 0.868                  & 0.622      & 0.240      & 0.346                  & 0.861      & 0.630      & \textbf{\underline{0.728}} & 0.144      & 0.129      & 0.137                  & 0.944      & 1.000      & 0.971                  & 0.891      & 0.932      & \textbf{\underline{0.911}}  \\
\textbf{FCVAE}               & 0.897      & 0.821      & \textbf{0.857}         & 0.897      & 0.792      & \textbf{0.842}         & 0.930      & 0.924      & \textbf{0.927}         & 0.906      & 0.772      & \textbf{0.835}         & 0.786      & 0.881      & \textbf{0.831}         & 0.705      & 0.571      & \textbf{0.631}         & 0.953      & 1.000      & \textbf{0.976}         & 0.925      & 0.909      & \textbf{0.917}          \\
\bottomrule
\end{tabular}
}
\end{table*}

Best F1 is obtained by traversing all possible thresholds for anomaly scores, and subsequently applying a point adjustment strategy to the prediction in order to compute the F1 score. Delay F1 is similar to best F1 but employs a delay point adjustment strategy to transform the prediction. The adjustment strategies are illustrated in Figure~\ref{pointadjust}, with a delay set to 1 as an example. The detector misses the second anomalous segment because it takes two-time intervals to detect this segment, exceeding the maximum delay threshold we established. We configure the delay for all datasets to be 7, except for Yahoo, where it is set to 3, and NAB, where it is set to 150. This is because the anomaly segments in Yahoo are very short, while in NAB, they are typically much longer, often spanning several hundred data points.

\subsubsection{Implementation Details}
To guarantee the widespread applicability, all the experiments described below were conducted under entirely unsupervised conditions, without employing any actual labels (all labels are set to zero). For consistency across all methods, we trained a single model for all curves within a dataset. Regarding hyperparameters, we conducted a grid search to identify the most effective parameters for different datasets. Additionally, we later evaluated the sensitivity of these parameters to ensure robust performance.

\subsection{Overall Performance}
The performance of FCVAE and baseline methods across the four datasets is depicted in Table \ref{bestf1}. Our method surpasses all baselines on four datasets regarding best F1 by 6.45\%, 0.98\%, 14.14\% and 0.31\%. In terms of delay F1, our method outperforms all baselines on four datasets by 4.98\%, 1.58\%, 38.68\% and 0.65\%.

The performance of various baseline methods on the datasets exhibits considerable variation. For instance, SPOT \cite{spot} does not excel on most datasets, as it erroneously treats extreme values as anomalies, whereas anomalies are not always manifested as such. SRCNN \cite{srcnn} is a reasonably proficient classifier, yet its performance falls short compared to most other models. This underscores the fact that implicitly extracting abnormal features is challenging. Informer \cite{informer} outperforms most other baselines across different datasets, as many anomalies exhibit notable value jumps, and prediction-based methods can effectively manage this situation. However, it struggles with anomalies induced by frequency changes. Anomaly-Transformer \cite{anomaly-transformer} attains commendable results on most datasets in terms of best F1 but demonstrates a low delay F1. It detects anomalies based on the relationships with nearby points, and only when the anomalous point is relatively central within the window can it easily capture the correlation. Conversely, TFAD \cite{tfad} achieves favorable results on various datasets but exhibits a certain delay in detection.

Moreover, our method surpasses DONUT \cite{donut} and VQRAE \cite{vqrae} in terms of reconstruction-based methods. Although VQRAE \cite{vqrae} introduces numerous modifications to the VAE, employing RNN to capture temporal relationships, our method still outperforms it. This finding implies that for UTS anomaly detection, it is imperative to incorporate only key information while avoiding overloading the model with superfluous data.

\subsection{Different Types of Conditions in CVAE}
We conduct experiments under identical settings to evaluate different types of conditions. The chosen conditions encompass information potentially useful for time series anomaly detection within the scope of our understanding, including timestamps \cite{anotransfer}, time domain information, and frequency domain information. To ensure consistency, we apply the same operation on the time domain information as we do on the frequency domain information.

\begin{figure*}[htbp]
\centering
\subfigure[Performance of CVAE using different conditions.\label{ablation:1}]{\includegraphics[scale=0.85]{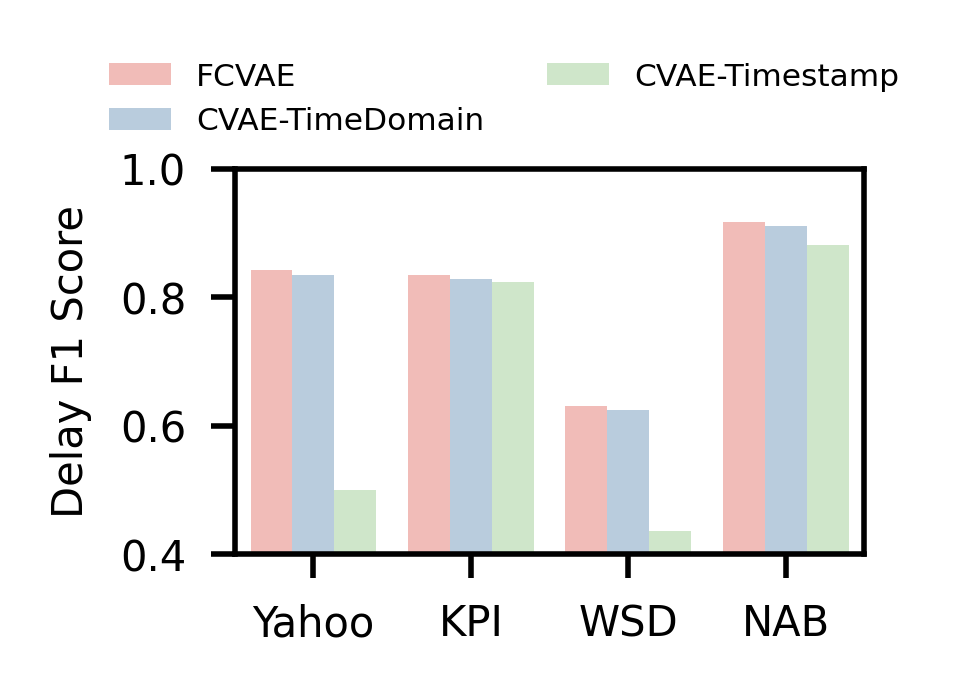}} \subfigure[Performance of different ways using frequency information.\label{ablation:2}]{\includegraphics[scale=0.85]{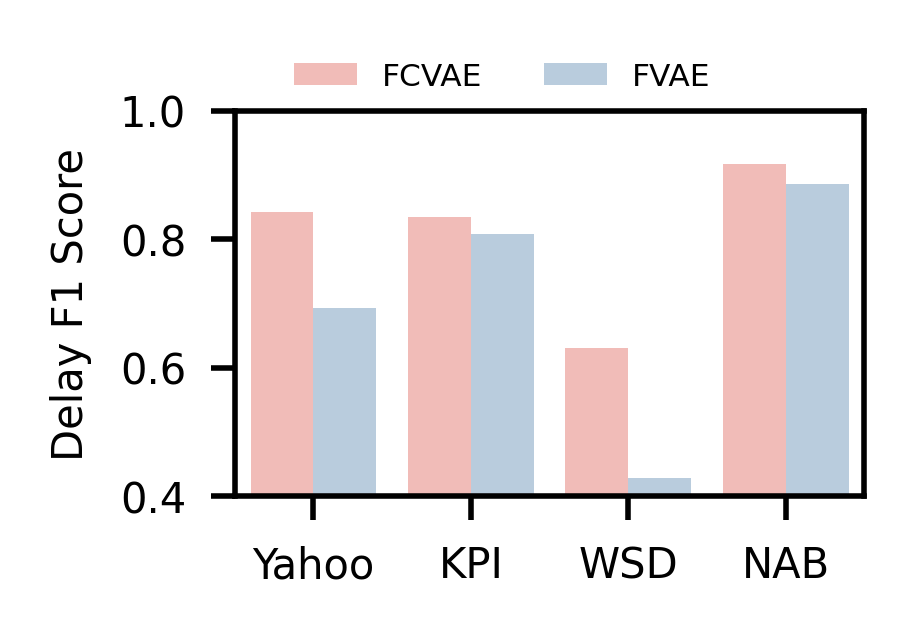}} \subfigure[Performance of different model structure.\label{ablation:3}]{\includegraphics[scale=0.85]{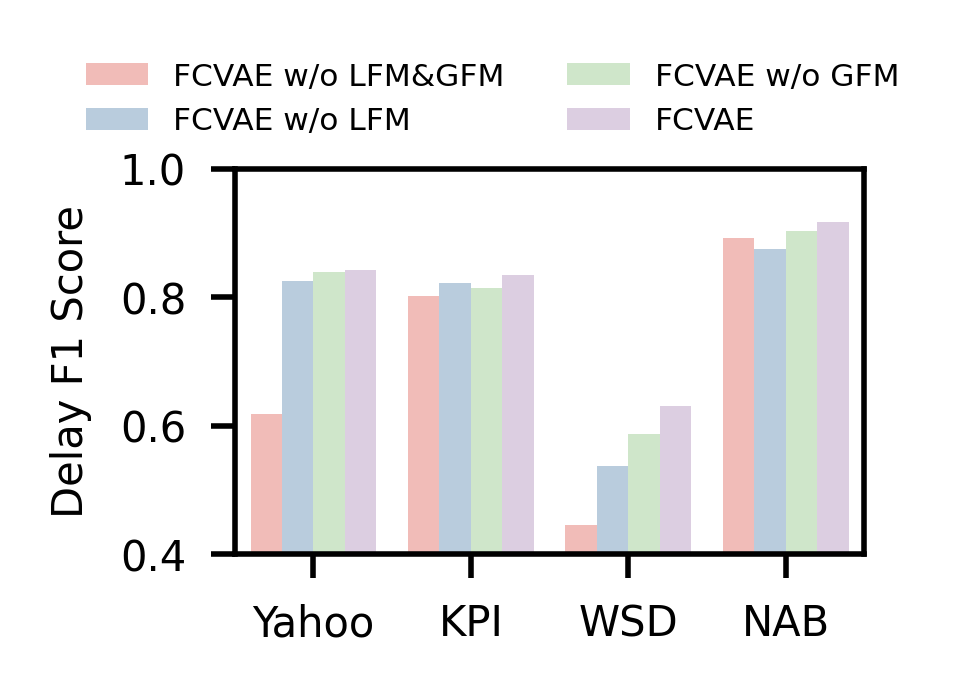}} \subfigure[Performance of whether using attention mechanism.\label{ablation:4}]{\includegraphics[scale=0.85]{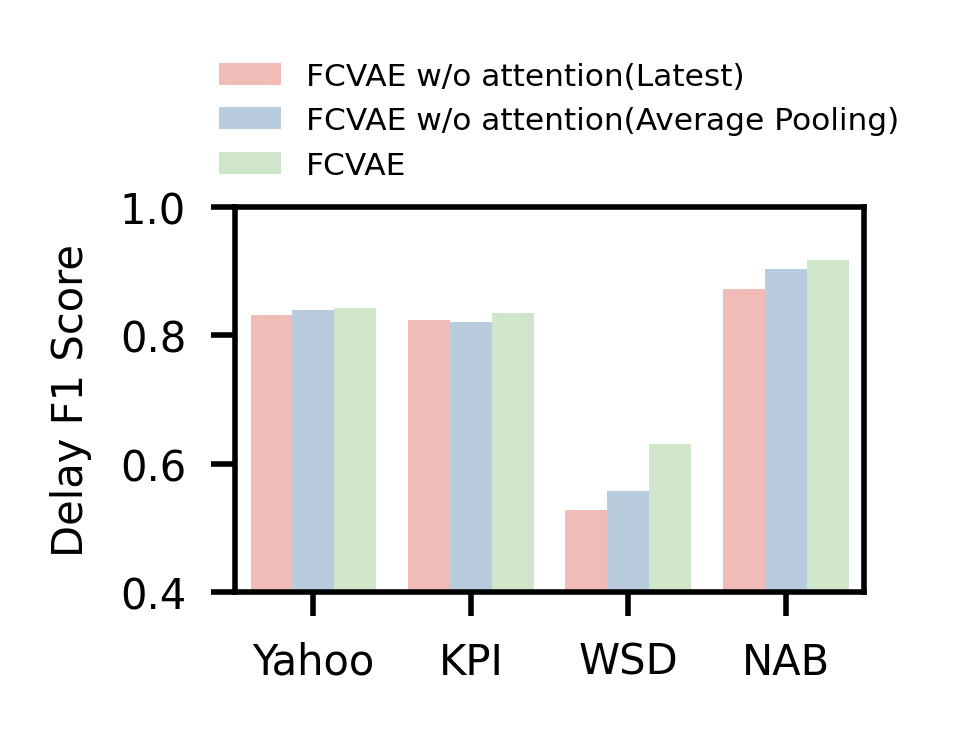}}
\caption{Delay F1 score of different settings.}
\label{ablation}
\end{figure*}

% \begin{figure}[htbp]
% \centerline{\includegraphics[scale=1]{pic/ab3.png}}
% \caption{Performance of CVAE using different conditions.}
% \label{ab3}
% \end{figure}

As illustrated in Figure~\ref{ablation:1}, the performance of employing the frequency information as a condition surpasses that using the timestamp or time domain information. This can be readily comprehended, as timestamps carry limited information and typically require one-hot encoding, resulting in sparse data representation. Time domain information is already incorporated in VAE, and utilizing it as a condition may lead to redundant information without significantly benefiting the reconstruction. Conversely, \textbf{frequency information, as a valuable and complementary prior, rendering it a more effective condition for anomaly detection.}

\subsection{Frequency VAE and FACVAE}
Is CVAE the optimal strategy for harnessing the frequency information in anomaly detection? In this study, we compare FCVAE with an improved frequency-based VAE (FVAE) model, in which the frequency information is integrated into VAE along with the input to reconstruct the original time series. As depicted in Figure~\ref{ablation:2}, FCVAE surpasses FVAE. This outcome can be attributed to two primary reasons. Firstly, CVAE, due to its unique architecture that incorporates conditional information, intrinsically outperforms VAE in numerous applications. Secondly, FVAE does not fully exploit frequency information. Although it incorporates this additional information, it still lacks efficient utilization in practice, particularly in the decoder. Consequently, \textbf{the CVAE that incorporates the frequency information as a condition represents the most effective structure known to date.}

% \begin{figure}[htbp]
% \centerline{\includegraphics[scale=1]{pic/ab4.png}}
% \caption{Performance of different ways using frequency information.}
% \label{ab4}
% \end{figure}

\subsection{GFM and LFM}
We propose GFM and LFM to extract global and local frequency information, respectively. However, do these two modules achieve our intended effects through their designs? Additionally, it is worth noting that GFM and LFM may overlap to some degree. Thus, we would like to determine if combining the two can further enhance the performance.

We conduct experiments and the results are depicted in Figure~\ref{ablation:3}. It can be observed that, across the four datasets, employing either LFM or GFM in FCVAE outperforms the VAE model under identical conditions of other settings except for NAB, where the frequent oscillation of data results in inconsistency between the information extracted from GFM and the data value of the current time. For all datasets, when both LFM and GFM modules are utilized concurrently, they synergistically enhance each other, resulting in superior performance. Consequently, \textbf{both global and local frequency information play a crucial role in detecting anomalies.}

% \begin{figure}[htbp]
% \centerline{\includegraphics[scale=1]{pic/ab2.png}}
% \caption{Performance of different model structure.}
% \label{ab2}
% \end{figure}

\subsection{Attention Mechanism}
It is crucial to discern whether the enhancement in LFM stems from the reduced window size or the attention mechanism. Thus, we perform experiments by excluding the attention operation from LFM while keeping GFM unaltered. Specifically, we utilized frequency information either from the latest small window in LFM (Latest) or from the average pooling of frequency information across all small windows in LFM (Average Pooling).

% \begin{figure}[htbp]
% \centerline{\includegraphics[scale=1]{pic/ab2_2.png}}
% \caption{Performance of whether using attention mechanism. The term ``Latest'' refers to using the frequency information from the most recent small window in LFM, while "Average Pooling" indicates utilizing the average of frequency information from all small windows within LFM.}
% \label{ab2_2}
% \end{figure}

The findings in Figure~\ref{ablation:4} demonstrate that without attention, it is impossible to attain the original performance of FCVAE since it is not feasible to determine the specific weight of each small window in advance. However, \textbf{the attention mechanism effectively addresses this issue by assigning higher weights to more informative windows.}

\begin{figure}[htbp]
\centering
\subfigure[Spectrum of small windows for data in the black dashed box on the right.\label{heatcase:a}]{\includegraphics[scale=0.42]{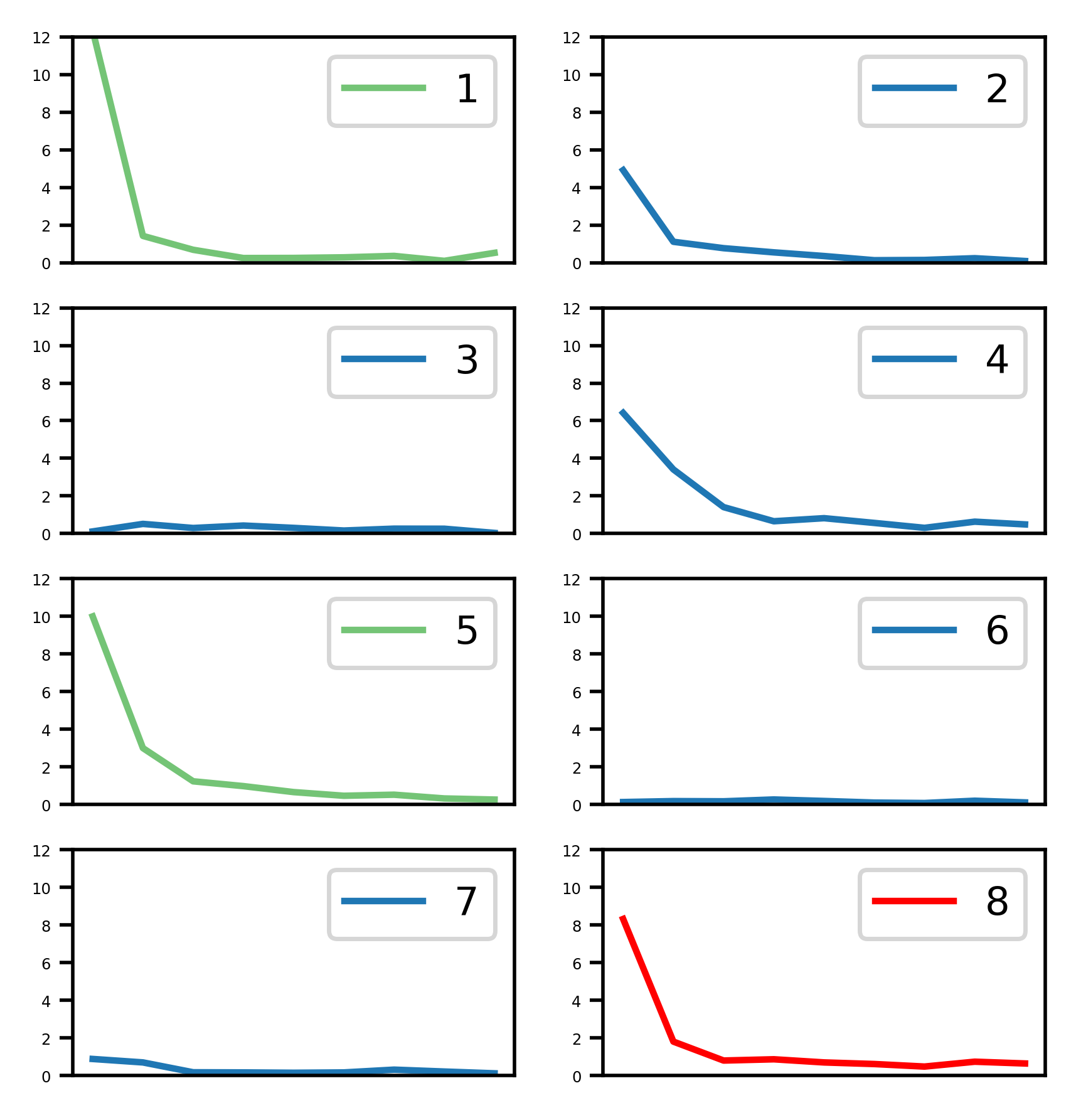}} \subfigure[Heatmap of LFM attention in a batch. The 8-th window is the latest window.\label{heatcase:b}]{\includegraphics[scale=0.42]{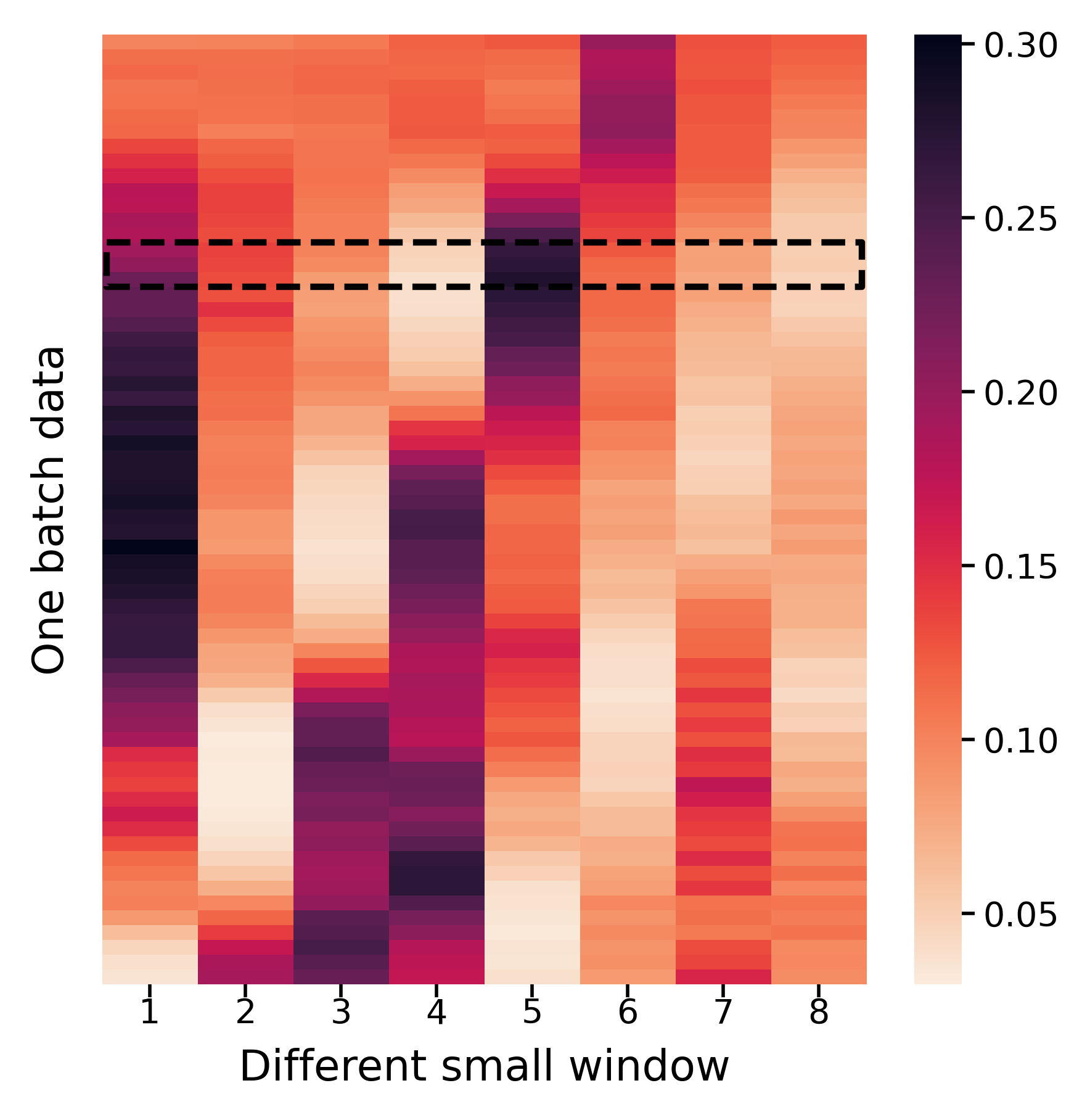}} 
\caption{An example of attention mechanism in LFM.}
\label{heatcase}
\end{figure}

\begin{figure*}[htbp]
\centering
\subfigure[Window Size]{\includegraphics[scale=0.8]{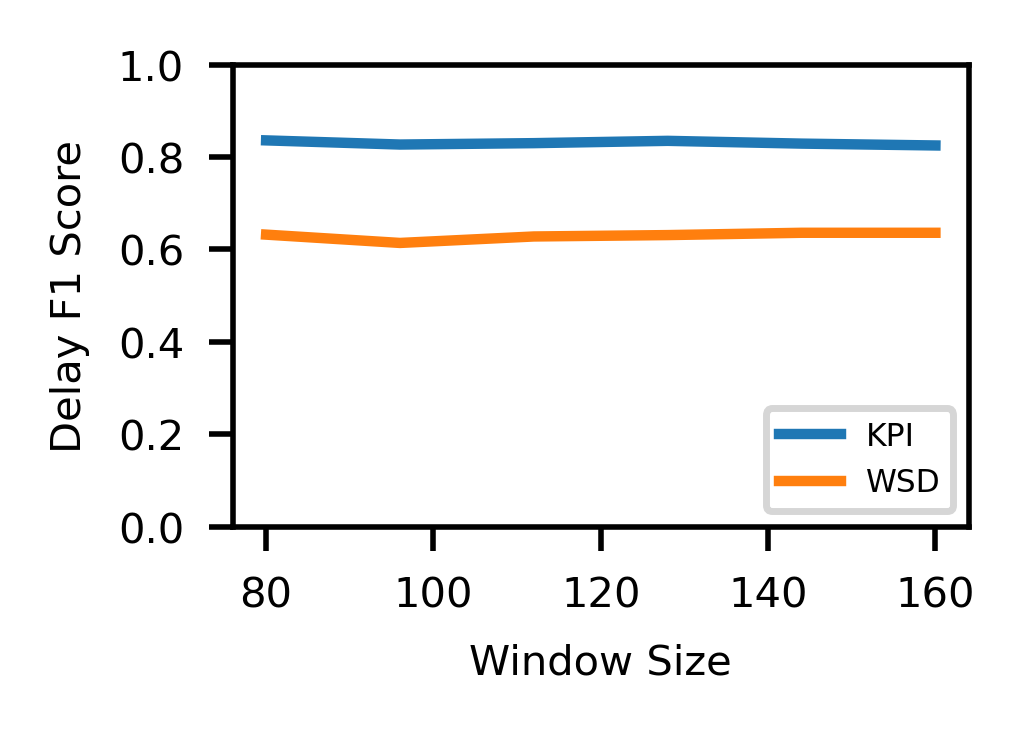}} \subfigure[Embedding Dimension]{\includegraphics[scale=0.8]{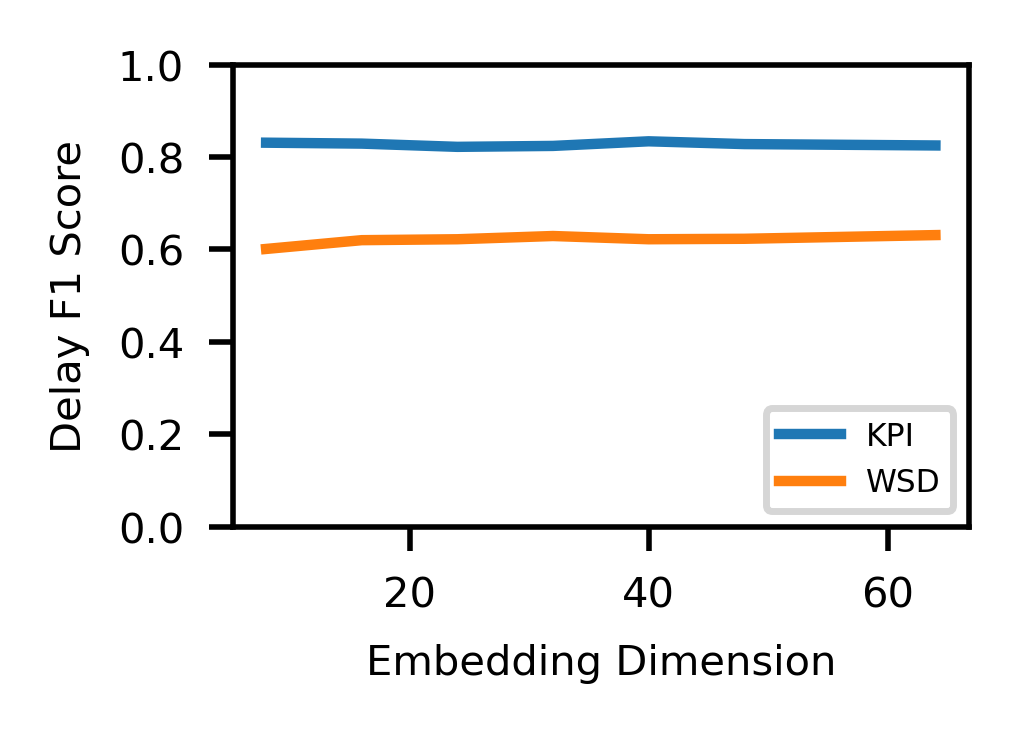}} \subfigure[Missing Data Injection Rate]{\includegraphics[scale=0.8]{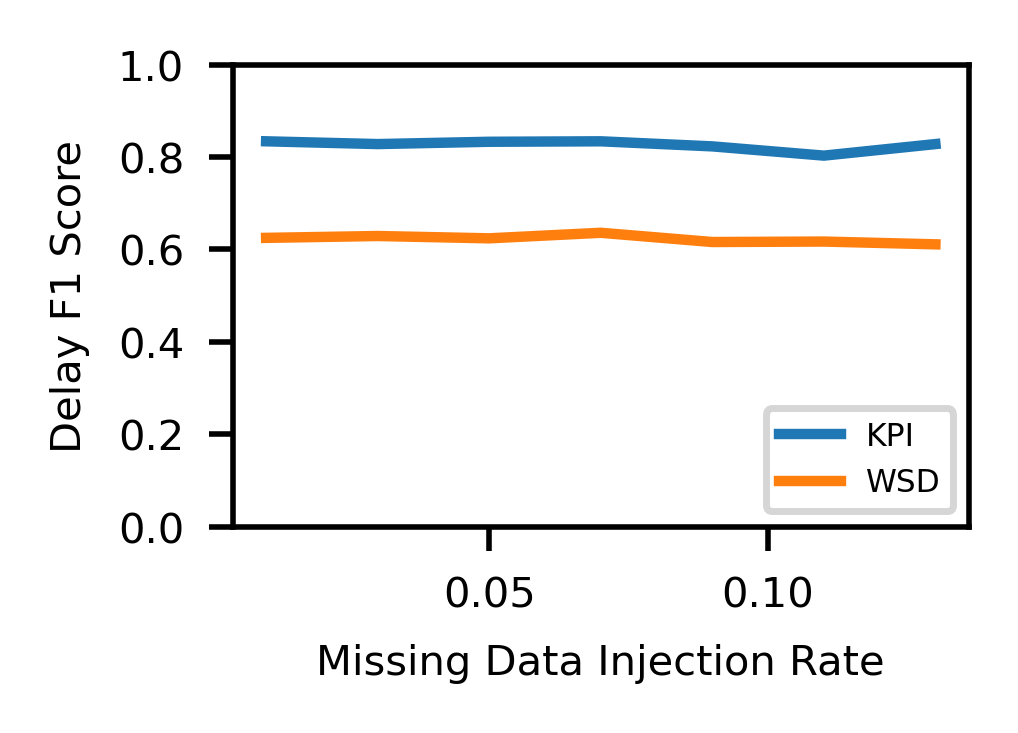}} \subfigure[Data Augment Rate]{\includegraphics[scale=0.8]{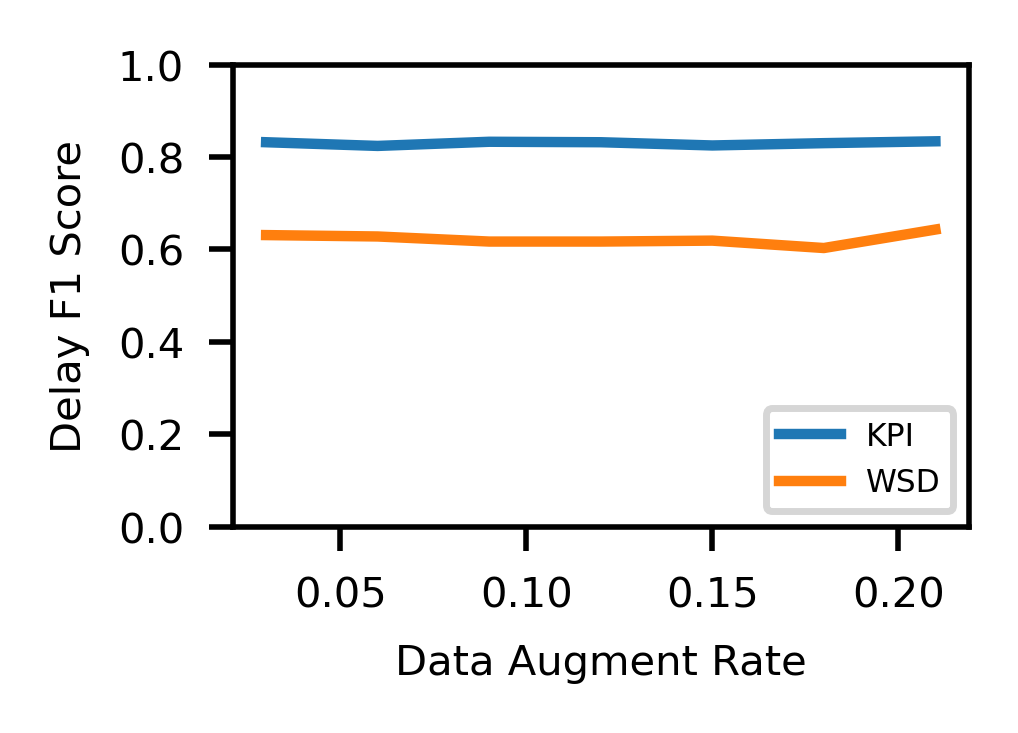}}
\caption{Delay F1 score of different settings.}
\label{sensi}
\end{figure*}

We present a comprehensive explanation of the attention mechanism in LFM using a case. A specific data segment, denoted by the black dashed box in Figure~\ref{heatcase:b}, is selected and all small windows produced by LFM's sliding window module are transformed into the frequency domain to obtain their spectra. As illustrated in Figure~\ref{heatcase:a}, the $5$-th (green) and the $8$-th (red) windows exhibit the highest similarity, where the $8$-th window serves as the query $Q$ for our attention. Upon examining Figure~\ref{heatcase:b}, it can be observed that the heat value of the $5$-th window is the highest, which corresponds with the findings in Figure~\ref{heatcase:a}. 

% Furthermore, we observe from Figure~\ref{heatcase:b} that the weight changes of LFM attention have a certain temporal gradient. This is easily understood since adjacent windows are similar.

\subsection{Key Techniques in Framework}
In this section, we evaluate the effectiveness of our novel data augmentation technique, masking the last point, and the application of CM-ELBO on four distinct datasets. The results are presented in Table \ref{tab-ablation1}. Based on the results, it is clear that CM-ELBO plays the most crucial role in most datasets, which aligns with our expectations. This is because it can tolerate abnormal or missing data to a certain extent. Furthermore, masking the last point has a substantial impact on the results, as when an anomaly occurs at the last point of the window, it affects the entire frequency information. Effectively masking this point resolves the issue and improves the detection accuracy. Data augmentation, on the other hand, introduces some artificial anomalies to boost the performance of CM-ELBO, particularly in unsupervised settings.

\begin{table}[htbp]
\caption{Delay F1 of different settings.}
\begin{center}
\small
\begin{tabular}{ccccc}
\toprule
Variants & Yahoo & KPI & WSD & NAB \\ \midrule
w/o data augment & 0.841&0.825&0.626&0.904 \\
w/o mask last point&0.835&0.830&0.534&0.877\\
w/o CM-ELBO &0.690&0.757&0.435&0.897\\
\textbf{FCVAE}&\textbf{0.842}&\textbf{0.835}&\textbf{0.631}&\textbf{0.917}\\
\bottomrule
\end{tabular}
\label{tab-ablation1}
\end{center}
\end{table}

\subsection{Parameter Sensitivity}
The stability of a model to different parameters is an important aspect to consider, and therefore we test the sensitivity of our model parameters on two datasets, KPI and WSD. We examine four aspects: the dimension of the condition, the window size, the proportion of missing data injection, and the proportion of data augmentation. The results, shown in Figure~\ref{sensi}, indicate that our model can achieve stable and excellent results under different parameter settings.%As a result, when applying our model to a new scenario, a simple parameter adjustment can be made to achieve a good F1 score, which is one of the advantages of our model.

\section{Production Impact and Efficiency}
Our FCVAE approach has been incorporated as a crucial component in a large-scale cloud system that caters to millions of users globally \cite{jiang2023xpert, chen2023empowering, jin2023assess}. The system generates billions of time series data points on a daily basis. 
The FCVAE detects anomalies in the cloud system, with the primary goal of identifying any potential regressions in the system that may indicate the occurrence of an incident. 

\begin{table}[htbp]
\caption{Online performance of FCVAE in production compared to legacy detector. F1 and F1$^*$ are defined in Table \ref{bestf1}.}
\begin{center}
\small
\begin{tabular}{ccccccc}
\toprule
\multicolumn{2}{c}{Baseline} &\multicolumn{2}{c}{FCVAE} & \multicolumn{2}{c}{Improvement} & Inference efficiency \\
F1 & F1$^*$& F1 & F1$^*$ & F1 &  F1$^*$ &[points/second]\\
\midrule
0.66&0.63 &0.73& 0.69& 10.9\%&11.1\%& 1195.7\\
\bottomrule
\end{tabular}
\label{online}
\end{center}
\end{table}

Table~\ref{online} presents the online performance improvement achieved by employing FCVAE over a period of one year. The experiments were conducted on a 24GB memory 3090 GPU. The results demonstrate substantial enhancements in both Best F1 and Delay F1 compared to the legacy detector. This underscores the effectiveness and robustness of our proposed method. Furthermore, our model is lightweight and highly efficient, capable of processing over 1000 data points within 1 second. This far exceeds the speed at which the system generates new temporal points.
% todo: write the results and inference efficiency (points/second) 

%We applied FCVAE to 22 online web systems of the top-tier company X, deploying a monitoring system for each system. Operators can register a task in the monitoring system, which extracts the UTS from the database and runs FCVAE. Ultimately, the monitoring results are visible on the monitoring dashboard.

\section{Related Work}
\textbf{Traditional statistical methods} \cite{ r1,r2,r3,r4,r5,r6,r7} are widely used in time series anomaly detection because of their great advantages in time series data processing. For example, \cite{r6} find the high frequency abnormal part of data through FFT\cite{r5} and verify it twice. Twitter\cite{r7} uses STL\cite{stl} to detect anomaly points. SPOT \cite{spot} considers that some extreme values are abnormal, therefore, detects them through Extreme Value Theory \cite{r9}.

\textbf{Supervised methods} \cite{r10,r11,srcnn, zhao2023robust} mostly learn the features of anomalies and identify them through classifiers based on the features learned. Opprentice \cite{r10} efficiently combines the results of many detectors through random forest. SRCNN \cite{srcnn} build a classifier through spectral residual \cite{sr} and CNN. Some methods \cite{r13,tfad} obtain pseudo-labels through data augmentation to enhance the learning ability.

\textbf{Unsupervised methods} are mainly divided into reconstruction-based and prediction-based methods. Reconstruction-based methods \cite{donut,begal,buzz,vqrae,contrast} learn low-dimensional representations and reconstruct the “normal patterns” of data and detect anomalies according to reconstruction error. DONUT\cite{donut} proposed the modified ELBO to enhance the capability of VAE in reconstructing the normal data. Buzz\cite{buzz} is the first to propose a deep generative model. ACVAE\cite{contrast} adds active learning and contrastive learning on the basis of VAE. Prediction-based methods \cite{lstm,informer} try to predict the normal values of metrics based on historical data and detect anomalies according to the prediction error. Informer\cite{informer} changes the relevant mechanism of self attention to achieve better prediction effect and efficiency. In recent years, transformer-based methods have been widely proposed. Anomaly-Transformer \cite{anomaly-transformer} detect anomalies by comparing Kullback-Leible (KL) divergence between two distributions. Some methods \cite{timesnet,fedformer} have begun to solve some practical problems from the frequency domain. Moerover, many transfer learning methods have been proposed\cite{ADT-SHL,ATAD,begal,anotransfer}. 

\section{Conclusion}
Our paper presents a novel unsupervised method for detecting anomalies in UTS, termed FCVAE. At the model level, we introduce the frequency domain information as a condition to work with CVAE. To capture the frequency information more accurately, we propose utilizing both GFM and LFM to concurrently capture the features from global and local frequency domains, and employing the target attention to more effectively extract local information. At the architecture level, we propose several new technologies, including CM-ELBO, data augmentation and masking the last point. We carry out experiments on four dataset and an online cloud system to evaluate our approach's accuracy, and comprehensive ablation experiments to demonstrate the effectiveness of each module.

\section{Acknowledgments}
% This work is supported by the National Key Research and Development Program of China (2021YFE0111500).
This work was supported in part by the National Key Research and Development Program of China (No.2021YFE0111500), in part by the National Natural Science Foundation of China (No.62202445), in part by the State Key Program of National Natural Science of China under Grant 62072264.

\bibliographystyle{ACM-Reference-Format}
\bibliography{fcvae}


\begin{thebibliography}{60}

%%% the Web address.

\ifx \showCODEN    \undefined \def \showCODEN     #1{\unskip}     \fi
\ifx \showDOI      \undefined \def \showDOI       #1{#1}\fi
\ifx \showISBNx    \undefined \def \showISBNx     #1{\unskip}     \fi
\ifx \showISBNxiii \undefined \def \showISBNxiii  #1{\unskip}     \fi
\ifx \showISSN     \undefined \def \showISSN      #1{\unskip}     \fi
\ifx \showLCCN     \undefined \def \showLCCN      #1{\unskip}     \fi
\ifx \shownote     \undefined \def \shownote      #1{#1}          \fi
\ifx \showarticletitle \undefined \def \showarticletitle #1{#1}   \fi
\ifx \showURL      \undefined \def \showURL       {\relax}        \fi
% invisible to TeX
\providecommand\bibfield[2]{#2}
\providecommand\bibinfo[2]{#2}
\providecommand\natexlab[1]{#1}
\providecommand\showeprint[2][]{arXiv:#2}

\bibitem[wsd({[n.\,d.]})]%
        {wsd2}
 \bibinfo{year}{[n.\,d.]}\natexlab{}.
\newblock \bibinfo{title}{WSD dataset}.
\newblock \bibinfo{howpublished}{Available: https://github.com/anotransfer/AnoTransfer-data/}.
\newblock


\bibitem[yah({[n.\,d.]})]%
        {yahoo}
 \bibinfo{year}{[n.\,d.]}\natexlab{}.
\newblock \bibinfo{title}{Yahoo dataset}.
\newblock \bibinfo{howpublished}{Available: https://webscope.sandbox.yahoo.com/}.
\newblock


\bibitem[Carmona et~al\mbox{.}(2021)]%
        {r13}
\bibfield{author}{\bibinfo{person}{Chris~U Carmona}, \bibinfo{person}{Fran{\c{c}}ois-Xavier Aubet}, \bibinfo{person}{Valentin Flunkert}, {and} \bibinfo{person}{Jan Gasthaus}.} \bibinfo{year}{2021}\natexlab{}.
\newblock \showarticletitle{Neural contextual anomaly detection for time series}.
\newblock \bibinfo{journal}{\emph{arXiv preprint arXiv:2107.07702}} (\bibinfo{year}{2021}).
\newblock


\bibitem[Chen et~al\mbox{.}(2021)]%
        {targetattention}
\bibfield{author}{\bibinfo{person}{Qiwei Chen}, \bibinfo{person}{Changhua Pei}, \bibinfo{person}{Shanshan Lv}, \bibinfo{person}{Chao Li}, \bibinfo{person}{Junfeng Ge}, {and} \bibinfo{person}{Wenwu Ou}.} \bibinfo{year}{2021}\natexlab{}.
\newblock \showarticletitle{End-to-end user behavior retrieval in click-through rateprediction model}.
\newblock \bibinfo{journal}{\emph{arXiv preprint arXiv:2108.04468}} (\bibinfo{year}{2021}).
\newblock


\bibitem[Chen et~al\mbox{.}(2019)]%
        {buzz}
\bibfield{author}{\bibinfo{person}{Wenxiao Chen}, \bibinfo{person}{Haowen Xu}, \bibinfo{person}{Zeyan Li}, \bibinfo{person}{Dan Pei}, \bibinfo{person}{Jie Chen}, \bibinfo{person}{Honglin Qiao}, \bibinfo{person}{Yang Feng}, {and} \bibinfo{person}{Zhaogang Wang}.} \bibinfo{year}{2019}\natexlab{}.
\newblock \showarticletitle{Unsupervised anomaly detection for intricate kpis via adversarial training of vae}. In \bibinfo{booktitle}{\emph{IEEE INFOCOM 2019-IEEE Conference on Computer Communications}}. IEEE, \bibinfo{pages}{1891--1899}.
\newblock


\bibitem[Chen et~al\mbox{.}(2024)]%
        {chen2023empowering}
\bibfield{author}{\bibinfo{person}{Yinfang Chen}, \bibinfo{person}{Huaibing Xie}, \bibinfo{person}{Minghua Ma}, \bibinfo{person}{Yu Kang}, \bibinfo{person}{Xin Gao}, \bibinfo{person}{Liu Shi}, \bibinfo{person}{Yunjie Cao}, \bibinfo{person}{Xuedong Gao}, \bibinfo{person}{Hao Fan}, \bibinfo{person}{Ming Wen}, {et~al\mbox{.}}} \bibinfo{year}{2024}\natexlab{}.
\newblock \showarticletitle{Automatic Root Cause Analysis via Large Language Models for Cloud Incidents}.
\newblock  (\bibinfo{year}{2024}).
\newblock


\bibitem[Chen et~al\mbox{.}(2023)]%
        {chen2023imdiffusion}
\bibfield{author}{\bibinfo{person}{Yuhang Chen}, \bibinfo{person}{Chaoyun Zhang}, \bibinfo{person}{Minghua Ma}, \bibinfo{person}{Yudong Liu}, \bibinfo{person}{Ruomeng Ding}, \bibinfo{person}{Bowen Li}, \bibinfo{person}{Shilin He}, \bibinfo{person}{Saravan Rajmohan}, \bibinfo{person}{Qingwei Lin}, {and} \bibinfo{person}{Dongmei Zhang}.} \bibinfo{year}{2023}\natexlab{}.
\newblock \showarticletitle{Imdiffusion: Imputed diffusion models for multivariate time series anomaly detection}.
\newblock \bibinfo{journal}{\emph{VLDB}} (\bibinfo{year}{2023}).
\newblock


\bibitem[Cleveland et~al\mbox{.}(1990)]%
        {stl}
\bibfield{author}{\bibinfo{person}{Robert~B Cleveland}, \bibinfo{person}{William~S Cleveland}, \bibinfo{person}{Jean~E McRae}, {and} \bibinfo{person}{Irma Terpenning}.} \bibinfo{year}{1990}\natexlab{}.
\newblock \showarticletitle{STL: A seasonal-trend decomposition}.
\newblock \bibinfo{journal}{\emph{J. Off. Stat}} \bibinfo{volume}{6}, \bibinfo{number}{1} (\bibinfo{year}{1990}), \bibinfo{pages}{3--73}.
\newblock


\bibitem[Dai et~al\mbox{.}(2021)]%
        {www5}
\bibfield{author}{\bibinfo{person}{Liang Dai}, \bibinfo{person}{Tao Lin}, \bibinfo{person}{Chang Liu}, \bibinfo{person}{Bo Jiang}, \bibinfo{person}{Yanwei Liu}, \bibinfo{person}{Zhen Xu}, {and} \bibinfo{person}{Zhi-Li Zhang}.} \bibinfo{year}{2021}\natexlab{}.
\newblock \showarticletitle{SDFVAE: Static and Dynamic Factorized VAE for Anomaly Detection of Multivariate CDN KPIs}. In \bibinfo{booktitle}{\emph{Proceedings of the Web Conference 2021}} (Ljubljana, Slovenia) \emph{(\bibinfo{series}{WWW '21})}. \bibinfo{publisher}{Association for Computing Machinery}, \bibinfo{address}{New York, NY, USA}, \bibinfo{pages}{3076–3086}.
\newblock
\showISBNx{9781450383127}
\urldef\tempurl%
\url{https://doi.org/10.1145/3442381.3450013}
\showDOI{\tempurl}


\bibitem[de~Haan and Ferreira(2006)]%
        {r9}
\bibfield{author}{\bibinfo{person}{L de Haan} {and} \bibinfo{person}{A Ferreira}.} \bibinfo{year}{2006}\natexlab{}.
\newblock \showarticletitle{Extreme Value Theory: an Introduction Springer Science+ Business Media}.
\newblock \bibinfo{journal}{\emph{LLC, New York}} (\bibinfo{year}{2006}).
\newblock


\bibitem[Deldari et~al\mbox{.}(2021)]%
        {www2}
\bibfield{author}{\bibinfo{person}{Shohreh Deldari}, \bibinfo{person}{Daniel~V. Smith}, \bibinfo{person}{Hao Xue}, {and} \bibinfo{person}{Flora~D. Salim}.} \bibinfo{year}{2021}\natexlab{}.
\newblock \showarticletitle{Time Series Change Point Detection with Self-Supervised Contrastive Predictive Coding}. In \bibinfo{booktitle}{\emph{Proceedings of the Web Conference 2021}} (Ljubljana, Slovenia) \emph{(\bibinfo{series}{WWW '21})}. \bibinfo{publisher}{Association for Computing Machinery}, \bibinfo{address}{New York, NY, USA}, \bibinfo{pages}{3124–3135}.
\newblock
\showISBNx{9781450383127}
\urldef\tempurl%
\url{https://doi.org/10.1145/3442381.3449903}
\showDOI{\tempurl}


\bibitem[Duan et~al\mbox{.}(2019)]%
        {ADT-SHL}
\bibfield{author}{\bibinfo{person}{XiaoYan Duan}, \bibinfo{person}{NingJiang Chen}, {and} \bibinfo{person}{YongSheng Xie}.} \bibinfo{year}{2019}\natexlab{}.
\newblock \showarticletitle{Intelligent detection of large-scale KPI streams anomaly based on transfer learning}. In \bibinfo{booktitle}{\emph{Big Data: 7th CCF Conference, BigData 2019, Wuhan, China, September 26--28, 2019, Proceedings 7}}. Springer, \bibinfo{pages}{366--379}.
\newblock


\bibitem[Ganatra et~al\mbox{.}(2023)]%
        {ganatra2023detection}
\bibfield{author}{\bibinfo{person}{Vaibhav Ganatra}, \bibinfo{person}{Anjaly Parayil}, \bibinfo{person}{Supriyo Ghosh}, \bibinfo{person}{Yu Kang}, \bibinfo{person}{Minghua Ma}, \bibinfo{person}{Chetan Bansal}, \bibinfo{person}{Suman Nath}, {and} \bibinfo{person}{Jonathan Mace}.} \bibinfo{year}{2023}\natexlab{}.
\newblock \showarticletitle{Detection Is Better Than Cure: A Cloud Incidents Perspective}. In \bibinfo{booktitle}{\emph{Proceedings of the 31st ACM Joint European Software Engineering Conference and Symposium on the Foundations of Software Engineering}}. \bibinfo{pages}{1891--1902}.
\newblock


\bibitem[G\"{u}nnemann et~al\mbox{.}(2014)]%
        {www7}
\bibfield{author}{\bibinfo{person}{Nikou G\"{u}nnemann}, \bibinfo{person}{Stephan G\"{u}nnemann}, {and} \bibinfo{person}{Christos Faloutsos}.} \bibinfo{year}{2014}\natexlab{}.
\newblock \showarticletitle{Robust Multivariate Autoregression for Anomaly Detection in Dynamic Product Ratings}. In \bibinfo{booktitle}{\emph{Proceedings of the 23rd International Conference on World Wide Web}} (Seoul, Korea) \emph{(\bibinfo{series}{WWW '14})}. \bibinfo{publisher}{Association for Computing Machinery}, \bibinfo{address}{New York, NY, USA}, \bibinfo{pages}{361–372}.
\newblock
\showISBNx{9781450327442}
\urldef\tempurl%
\url{https://doi.org/10.1145/2566486.2568008}
\showDOI{\tempurl}


\bibitem[Hou and Zhang(2007)]%
        {sr}
\bibfield{author}{\bibinfo{person}{Xiaodi Hou} {and} \bibinfo{person}{Liqing Zhang}.} \bibinfo{year}{2007}\natexlab{}.
\newblock \showarticletitle{Saliency detection: A spectral residual approach}. In \bibinfo{booktitle}{\emph{2007 IEEE Conference on computer vision and pattern recognition}}. Ieee, \bibinfo{pages}{1--8}.
\newblock


\bibitem[Huang et~al\mbox{.}(2019)]%
        {dsanet}
\bibfield{author}{\bibinfo{person}{Siteng Huang}, \bibinfo{person}{Donglin Wang}, \bibinfo{person}{Xuehan Wu}, {and} \bibinfo{person}{Ao Tang}.} \bibinfo{year}{2019}\natexlab{}.
\newblock \showarticletitle{Dsanet: Dual self-attention network for multivariate time series forecasting}. In \bibinfo{booktitle}{\emph{Proceedings of the 28th ACM international conference on information and knowledge management}}. \bibinfo{pages}{2129--2132}.
\newblock


\bibitem[Huang et~al\mbox{.}(2022)]%
        {www1}
\bibfield{author}{\bibinfo{person}{Tao Huang}, \bibinfo{person}{Pengfei Chen}, {and} \bibinfo{person}{Ruipeng Li}.} \bibinfo{year}{2022}\natexlab{}.
\newblock \showarticletitle{A Semi-Supervised VAE Based Active Anomaly Detection Framework in Multivariate Time Series for Online Systems}. In \bibinfo{booktitle}{\emph{Proceedings of the ACM Web Conference 2022}} (Virtual Event, Lyon, France) \emph{(\bibinfo{series}{WWW '22})}. \bibinfo{publisher}{Association for Computing Machinery}, \bibinfo{address}{New York, NY, USA}, \bibinfo{pages}{1797–1806}.
\newblock
\showISBNx{9781450390965}
\urldef\tempurl%
\url{https://doi.org/10.1145/3485447.3511984}
\showDOI{\tempurl}


\bibitem[Hundman et~al\mbox{.}(2018)]%
        {lstm}
\bibfield{author}{\bibinfo{person}{Kyle Hundman}, \bibinfo{person}{Valentino Constantinou}, \bibinfo{person}{Christopher Laporte}, \bibinfo{person}{Ian Colwell}, {and} \bibinfo{person}{Tom Soderstrom}.} \bibinfo{year}{2018}\natexlab{}.
\newblock \showarticletitle{Detecting spacecraft anomalies using lstms and nonparametric dynamic thresholding}. In \bibinfo{booktitle}{\emph{Proceedings of the 24th ACM SIGKDD international conference on knowledge discovery \& data mining}}. \bibinfo{pages}{387--395}.
\newblock


\bibitem[Jiang et~al\mbox{.}(2024)]%
        {jiang2023xpert}
\bibfield{author}{\bibinfo{person}{Yuxuan Jiang}, \bibinfo{person}{Chaoyun Zhang}, \bibinfo{person}{Shilin He}, \bibinfo{person}{Zhihao Yang}, \bibinfo{person}{Minghua Ma}, \bibinfo{person}{Si Qin}, \bibinfo{person}{Yu Kang}, \bibinfo{person}{Yingnong Dang}, \bibinfo{person}{Saravan Rajmohan}, \bibinfo{person}{Qingwei Lin}, {et~al\mbox{.}}} \bibinfo{year}{2024}\natexlab{}.
\newblock \showarticletitle{Xpert: Empowering Incident Management with Query Recommendations via Large Language Models}.
\newblock  (\bibinfo{year}{2024}).
\newblock


\bibitem[Jin et~al\mbox{.}(2023)]%
        {jin2023assess}
\bibfield{author}{\bibinfo{person}{Pengxiang Jin}, \bibinfo{person}{Shenglin Zhang}, \bibinfo{person}{Minghua Ma}, \bibinfo{person}{Haozhe Li}, \bibinfo{person}{Yu Kang}, \bibinfo{person}{Liqun Li}, \bibinfo{person}{Yudong Liu}, \bibinfo{person}{Bo Qiao}, \bibinfo{person}{Chaoyun Zhang}, \bibinfo{person}{Pu Zhao}, {et~al\mbox{.}}} \bibinfo{year}{2023}\natexlab{}.
\newblock \showarticletitle{Assess and Summarize: Improve Outage Understanding with Large Language Models}.
\newblock  (\bibinfo{year}{2023}).
\newblock


\bibitem[Kamarthi et~al\mbox{.}(2022)]%
        {www8}
\bibfield{author}{\bibinfo{person}{Harshavardhan Kamarthi}, \bibinfo{person}{Lingkai Kong}, \bibinfo{person}{Alexander Rodriguez}, \bibinfo{person}{Chao Zhang}, {and} \bibinfo{person}{B~Aditya Prakash}.} \bibinfo{year}{2022}\natexlab{}.
\newblock \showarticletitle{CAMul: Calibrated and Accurate Multi-View Time-Series Forecasting}. In \bibinfo{booktitle}{\emph{Proceedings of the ACM Web Conference 2022}} (Virtual Event, Lyon, France) \emph{(\bibinfo{series}{WWW '22})}. \bibinfo{publisher}{Association for Computing Machinery}, \bibinfo{address}{New York, NY, USA}, \bibinfo{pages}{3174–3185}.
\newblock
\showISBNx{9781450390965}
\urldef\tempurl%
\url{https://doi.org/10.1145/3485447.3512037}
\showDOI{\tempurl}


\bibitem[Kieu et~al\mbox{.}(2022)]%
        {vqrae}
\bibfield{author}{\bibinfo{person}{Tung Kieu}, \bibinfo{person}{Bin Yang}, \bibinfo{person}{Chenjuan Guo}, \bibinfo{person}{Razvan-Gabriel Cirstea}, \bibinfo{person}{Yan Zhao}, \bibinfo{person}{Yale Song}, {and} \bibinfo{person}{Christian~S Jensen}.} \bibinfo{year}{2022}\natexlab{}.
\newblock \showarticletitle{Anomaly detection in time series with robust variational quasi-recurrent autoencoders}. In \bibinfo{booktitle}{\emph{2022 IEEE 38th International Conference on Data Engineering (ICDE)}}. IEEE, \bibinfo{pages}{1342--1354}.
\newblock


\bibitem[Kingma and Welling(2013)]%
        {vae}
\bibfield{author}{\bibinfo{person}{Diederik~P Kingma} {and} \bibinfo{person}{Max Welling}.} \bibinfo{year}{2013}\natexlab{}.
\newblock \showarticletitle{Auto-encoding variational bayes}.
\newblock \bibinfo{journal}{\emph{arXiv preprint arXiv:1312.6114}} (\bibinfo{year}{2013}).
\newblock


\bibitem[Laptev et~al\mbox{.}(2015)]%
        {r11}
\bibfield{author}{\bibinfo{person}{Nikolay Laptev}, \bibinfo{person}{Saeed Amizadeh}, {and} \bibinfo{person}{Ian Flint}.} \bibinfo{year}{2015}\natexlab{}.
\newblock \showarticletitle{Generic and scalable framework for automated time-series anomaly detection}. In \bibinfo{booktitle}{\emph{Proceedings of the 21th ACM SIGKDD international conference on knowledge discovery and data mining}}. \bibinfo{pages}{1939--1947}.
\newblock


\bibitem[Lavin and Ahmad(2015)]%
        {nab}
\bibfield{author}{\bibinfo{person}{Alexander Lavin} {and} \bibinfo{person}{Subutai Ahmad}.} \bibinfo{year}{2015}\natexlab{}.
\newblock \showarticletitle{Evaluating real-time anomaly detection algorithms--the Numenta anomaly benchmark}. In \bibinfo{booktitle}{\emph{2015 IEEE 14th international conference on machine learning and applications (ICMLA)}}. IEEE, \bibinfo{pages}{38--44}.
\newblock


\bibitem[Le~Guennec et~al\mbox{.}(2016)]%
        {dataaug1}
\bibfield{author}{\bibinfo{person}{Arthur Le~Guennec}, \bibinfo{person}{Simon Malinowski}, {and} \bibinfo{person}{Romain Tavenard}.} \bibinfo{year}{2016}\natexlab{}.
\newblock \showarticletitle{Data augmentation for time series classification using convolutional neural networks}. In \bibinfo{booktitle}{\emph{ECML/PKDD workshop on advanced analytics and learning on temporal data}}.
\newblock


\bibitem[Li et~al\mbox{.}(2018)]%
        {begal}
\bibfield{author}{\bibinfo{person}{Zeyan Li}, \bibinfo{person}{Wenxiao Chen}, {and} \bibinfo{person}{Dan Pei}.} \bibinfo{year}{2018}\natexlab{}.
\newblock \showarticletitle{Robust and unsupervised kpi anomaly detection based on conditional variational autoencoder}. In \bibinfo{booktitle}{\emph{2018 IEEE 37th International Performance Computing and Communications Conference (IPCCC)}}. IEEE, \bibinfo{pages}{1--9}.
\newblock


\bibitem[Li et~al\mbox{.}(2022b)]%
        {aiops}
\bibfield{author}{\bibinfo{person}{Zeyan Li}, \bibinfo{person}{Nengwen Zhao}, \bibinfo{person}{Shenglin Zhang}, \bibinfo{person}{Yongqian Sun}, \bibinfo{person}{Pengfei Chen}, \bibinfo{person}{Xidao Wen}, \bibinfo{person}{Minghua Ma}, {and} \bibinfo{person}{Dan Pei}.} \bibinfo{year}{2022}\natexlab{b}.
\newblock \showarticletitle{Constructing Large-Scale Real-World Benchmark Datasets for AIOps}.
\newblock \bibinfo{journal}{\emph{arXiv preprint arXiv:2208.03938}} (\bibinfo{year}{2022}).
\newblock


\bibitem[Li et~al\mbox{.}(2022a)]%
        {contrast}
\bibfield{author}{\bibinfo{person}{Zhihan Li}, \bibinfo{person}{Youjian Zhao}, \bibinfo{person}{Yitong Geng}, \bibinfo{person}{Zhanxiang Zhao}, \bibinfo{person}{Hanzhang Wang}, \bibinfo{person}{Wenxiao Chen}, \bibinfo{person}{Huai Jiang}, \bibinfo{person}{Amber Vaidya}, \bibinfo{person}{Liangfei Su}, {and} \bibinfo{person}{Dan Pei}.} \bibinfo{year}{2022}\natexlab{a}.
\newblock \showarticletitle{Situation-Aware Multivariate Time Series Anomaly Detection Through Active Learning and Contrast VAE-Based Models in Large Distributed Systems}.
\newblock \bibinfo{journal}{\emph{IEEE Journal on Selected Areas in Communications}} \bibinfo{volume}{40}, \bibinfo{number}{9} (\bibinfo{year}{2022}), \bibinfo{pages}{2746--2765}.
\newblock


\bibitem[Li et~al\mbox{.}(2021)]%
        {interfusion}
\bibfield{author}{\bibinfo{person}{Zhihan Li}, \bibinfo{person}{Youjian Zhao}, \bibinfo{person}{Jiaqi Han}, \bibinfo{person}{Ya Su}, \bibinfo{person}{Rui Jiao}, \bibinfo{person}{Xidao Wen}, {and} \bibinfo{person}{Dan Pei}.} \bibinfo{year}{2021}\natexlab{}.
\newblock \showarticletitle{Multivariate time series anomaly detection and interpretation using hierarchical inter-metric and temporal embedding}. In \bibinfo{booktitle}{\emph{Proceedings of the 27th ACM SIGKDD conference on knowledge discovery \& data mining}}. \bibinfo{pages}{3220--3230}.
\newblock


\bibitem[Liu et~al\mbox{.}(2015)]%
        {r10}
\bibfield{author}{\bibinfo{person}{Dapeng Liu}, \bibinfo{person}{Youjian Zhao}, \bibinfo{person}{Haowen Xu}, \bibinfo{person}{Yongqian Sun}, \bibinfo{person}{Dan Pei}, \bibinfo{person}{Jiao Luo}, \bibinfo{person}{Xiaowei Jing}, {and} \bibinfo{person}{Mei Feng}.} \bibinfo{year}{2015}\natexlab{}.
\newblock \showarticletitle{Opprentice: Towards practical and automatic anomaly detection through machine learning}. In \bibinfo{booktitle}{\emph{Proceedings of the 2015 internet measurement conference}}. \bibinfo{pages}{211--224}.
\newblock


\bibitem[Lu and Ghorbani(2008)]%
        {r2}
\bibfield{author}{\bibinfo{person}{Wei Lu} {and} \bibinfo{person}{Ali~A Ghorbani}.} \bibinfo{year}{2008}\natexlab{}.
\newblock \showarticletitle{Network anomaly detection based on wavelet analysis}.
\newblock \bibinfo{journal}{\emph{EURASIP Journal on Advances in Signal processing}}  \bibinfo{volume}{2009} (\bibinfo{year}{2008}), \bibinfo{pages}{1--16}.
\newblock


\bibitem[Lu et~al\mbox{.}(2023)]%
        {www6}
\bibfield{author}{\bibinfo{person}{Xiaofeng Lu}, \bibinfo{person}{Xiaoyu Zhang}, {and} \bibinfo{person}{Pietro Lio}.} \bibinfo{year}{2023}\natexlab{}.
\newblock \showarticletitle{GAT-DNS: DNS Multivariate Time Series Prediction Model Based on Graph Attention Network}. In \bibinfo{booktitle}{\emph{Companion Proceedings of the ACM Web Conference 2023}} (Austin, TX, USA) \emph{(\bibinfo{series}{WWW '23 Companion})}. \bibinfo{publisher}{Association for Computing Machinery}, \bibinfo{address}{New York, NY, USA}, \bibinfo{pages}{127–131}.
\newblock
\showISBNx{9781450394192}
\urldef\tempurl%
\url{https://doi.org/10.1145/3543873.3587329}
\showDOI{\tempurl}


\bibitem[Ma et~al\mbox{.}(2021)]%
        {ma2021jump}
\bibfield{author}{\bibinfo{person}{Minghua Ma}, \bibinfo{person}{Shenglin Zhang}, \bibinfo{person}{Junjie Chen}, \bibinfo{person}{Jim Xu}, \bibinfo{person}{Haozhe Li}, \bibinfo{person}{Yongliang Lin}, \bibinfo{person}{Xiaohui Nie}, \bibinfo{person}{Bo Zhou}, \bibinfo{person}{Yong Wang}, {and} \bibinfo{person}{Dan Pei}.} \bibinfo{year}{2021}\natexlab{}.
\newblock \showarticletitle{$\{$Jump-Starting$\}$ Multivariate Time Series Anomaly Detection for Online Service Systems}. In \bibinfo{booktitle}{\emph{2021 USENIX Annual Technical Conference (USENIX ATC 21)}}. \bibinfo{pages}{413--426}.
\newblock


\bibitem[Mahimkar et~al\mbox{.}(2011)]%
        {r3}
\bibfield{author}{\bibinfo{person}{Ajay Mahimkar}, \bibinfo{person}{Zihui Ge}, \bibinfo{person}{Jia Wang}, \bibinfo{person}{Jennifer Yates}, \bibinfo{person}{Yin Zhang}, \bibinfo{person}{Joanne Emmons}, \bibinfo{person}{Brian Huntley}, {and} \bibinfo{person}{Mark Stockert}.} \bibinfo{year}{2011}\natexlab{}.
\newblock \showarticletitle{Rapid detection of maintenance induced changes in service performance}. In \bibinfo{booktitle}{\emph{Proceedings of the Seventh COnference on Emerging Networking EXperiments and Technologies}}. \bibinfo{pages}{1--12}.
\newblock


\bibitem[Ovadia et~al\mbox{.}(2022)]%
        {www4}
\bibfield{author}{\bibinfo{person}{Oded Ovadia}, \bibinfo{person}{Oren Elisha}, {and} \bibinfo{person}{Elad Yom-Tov}.} \bibinfo{year}{2022}\natexlab{}.
\newblock \showarticletitle{Detection of Infectious Disease Outbreaks in Search Engine Time Series Using Non-Specific Syndromic Surveillance with Effect-Size Filtering}. In \bibinfo{booktitle}{\emph{Companion Proceedings of the Web Conference 2022}} (Virtual Event, Lyon, France) \emph{(\bibinfo{series}{WWW '22})}. \bibinfo{publisher}{Association for Computing Machinery}, \bibinfo{address}{New York, NY, USA}, \bibinfo{pages}{924–929}.
\newblock
\showISBNx{9781450391306}
\urldef\tempurl%
\url{https://doi.org/10.1145/3487553.3524672}
\showDOI{\tempurl}


\bibitem[Rasheed et~al\mbox{.}(2009)]%
        {r6}
\bibfield{author}{\bibinfo{person}{Faraz Rasheed}, \bibinfo{person}{Peter Peng}, \bibinfo{person}{Reda Alhajj}, {and} \bibinfo{person}{Jon Rokne}.} \bibinfo{year}{2009}\natexlab{}.
\newblock \showarticletitle{Fourier transform based spatial outlier mining}. In \bibinfo{booktitle}{\emph{Intelligent Data Engineering and Automated Learning-IDEAL 2009: 10th International Conference, Burgos, Spain, September 23-26, 2009. Proceedings 10}}. Springer, \bibinfo{pages}{317--324}.
\newblock


\bibitem[Ren et~al\mbox{.}(2019)]%
        {srcnn}
\bibfield{author}{\bibinfo{person}{Hansheng Ren}, \bibinfo{person}{Bixiong Xu}, \bibinfo{person}{Yujing Wang}, \bibinfo{person}{Chao Yi}, \bibinfo{person}{Congrui Huang}, \bibinfo{person}{Xiaoyu Kou}, \bibinfo{person}{Tony Xing}, \bibinfo{person}{Mao Yang}, \bibinfo{person}{Jie Tong}, {and} \bibinfo{person}{Qi Zhang}.} \bibinfo{year}{2019}\natexlab{}.
\newblock \showarticletitle{Time-series anomaly detection service at microsoft}. In \bibinfo{booktitle}{\emph{Proceedings of the 25th ACM SIGKDD international conference on knowledge discovery \& data mining}}. \bibinfo{pages}{3009--3017}.
\newblock


\bibitem[Rezende et~al\mbox{.}(2014)]%
        {stochastic}
\bibfield{author}{\bibinfo{person}{Danilo~Jimenez Rezende}, \bibinfo{person}{Shakir Mohamed}, {and} \bibinfo{person}{Daan Wierstra}.} \bibinfo{year}{2014}\natexlab{}.
\newblock \showarticletitle{Stochastic backpropagation and approximate inference in deep generative models}. In \bibinfo{booktitle}{\emph{International conference on machine learning}}. PMLR, \bibinfo{pages}{1278--1286}.
\newblock


\bibitem[Rosner(1983)]%
        {r1}
\bibfield{author}{\bibinfo{person}{Bernard Rosner}.} \bibinfo{year}{1983}\natexlab{}.
\newblock \showarticletitle{Percentage points for a generalized ESD many-outlier procedure}.
\newblock \bibinfo{journal}{\emph{Technometrics}} \bibinfo{volume}{25}, \bibinfo{number}{2} (\bibinfo{year}{1983}), \bibinfo{pages}{165--172}.
\newblock


\bibitem[Shen et~al\mbox{.}(2020)]%
        {rare}
\bibfield{author}{\bibinfo{person}{Lifeng Shen}, \bibinfo{person}{Zhuocong Li}, {and} \bibinfo{person}{James Kwok}.} \bibinfo{year}{2020}\natexlab{}.
\newblock \showarticletitle{Timeseries anomaly detection using temporal hierarchical one-class network}.
\newblock \bibinfo{journal}{\emph{Advances in Neural Information Processing Systems}}  \bibinfo{volume}{33} (\bibinfo{year}{2020}), \bibinfo{pages}{13016--13026}.
\newblock


\bibitem[Siffer et~al\mbox{.}(2017)]%
        {spot}
\bibfield{author}{\bibinfo{person}{Alban Siffer}, \bibinfo{person}{Pierre-Alain Fouque}, \bibinfo{person}{Alexandre Termier}, {and} \bibinfo{person}{Christine Largouet}.} \bibinfo{year}{2017}\natexlab{}.
\newblock \showarticletitle{Anomaly detection in streams with extreme value theory}. In \bibinfo{booktitle}{\emph{Proceedings of the 23rd ACM SIGKDD International Conference on Knowledge Discovery and Data Mining}}. \bibinfo{pages}{1067--1075}.
\newblock


\bibitem[Sohn et~al\mbox{.}(2015)]%
        {cvae}
\bibfield{author}{\bibinfo{person}{Kihyuk Sohn}, \bibinfo{person}{Honglak Lee}, {and} \bibinfo{person}{Xinchen Yan}.} \bibinfo{year}{2015}\natexlab{}.
\newblock \showarticletitle{Learning structured output representation using deep conditional generative models}.
\newblock \bibinfo{journal}{\emph{Advances in neural information processing systems}}  \bibinfo{volume}{28} (\bibinfo{year}{2015}).
\newblock


\bibitem[Vallis et~al\mbox{.}(2014)]%
        {r7}
\bibfield{author}{\bibinfo{person}{Owen Vallis}, \bibinfo{person}{Jordan Hochenbaum}, {and} \bibinfo{person}{Arun Kejariwal}.} \bibinfo{year}{2014}\natexlab{}.
\newblock \showarticletitle{A novel technique for long-term anomaly detection in the cloud}. In \bibinfo{booktitle}{\emph{6th $\{$USENIX$\}$ workshop on hot topics in cloud computing (HotCloud 14)}}.
\newblock


\bibitem[Van~Loan(1992)]%
        {r5}
\bibfield{author}{\bibinfo{person}{Charles Van~Loan}.} \bibinfo{year}{1992}\natexlab{}.
\newblock \bibinfo{booktitle}{\emph{Computational frameworks for the fast Fourier transform}}.
\newblock \bibinfo{publisher}{SIAM}.
\newblock


\bibitem[Vaswani et~al\mbox{.}(2017)]%
        {attention}
\bibfield{author}{\bibinfo{person}{Ashish Vaswani}, \bibinfo{person}{Noam Shazeer}, \bibinfo{person}{Niki Parmar}, \bibinfo{person}{Jakob Uszkoreit}, \bibinfo{person}{Llion Jones}, \bibinfo{person}{Aidan~N Gomez}, \bibinfo{person}{{\L}ukasz Kaiser}, {and} \bibinfo{person}{Illia Polosukhin}.} \bibinfo{year}{2017}\natexlab{}.
\newblock \showarticletitle{Attention is all you need}.
\newblock \bibinfo{journal}{\emph{Advances in neural information processing systems}}  \bibinfo{volume}{30} (\bibinfo{year}{2017}).
\newblock


\bibitem[Wen et~al\mbox{.}(2020)]%
        {dataaug2}
\bibfield{author}{\bibinfo{person}{Qingsong Wen}, \bibinfo{person}{Liang Sun}, \bibinfo{person}{Fan Yang}, \bibinfo{person}{Xiaomin Song}, \bibinfo{person}{Jingkun Gao}, \bibinfo{person}{Xue Wang}, {and} \bibinfo{person}{Huan Xu}.} \bibinfo{year}{2020}\natexlab{}.
\newblock \showarticletitle{Time series data augmentation for deep learning: A survey}.
\newblock \bibinfo{journal}{\emph{arXiv preprint arXiv:2002.12478}} (\bibinfo{year}{2020}).
\newblock


\bibitem[Wu et~al\mbox{.}(2023)]%
        {timesnet}
\bibfield{author}{\bibinfo{person}{Haixu Wu}, \bibinfo{person}{Tengge Hu}, \bibinfo{person}{Yong Liu}, \bibinfo{person}{Hang Zhou}, \bibinfo{person}{Jianmin Wang}, {and} \bibinfo{person}{Mingsheng Long}.} \bibinfo{year}{2023}\natexlab{}.
\newblock \showarticletitle{TimesNet: Temporal 2D-Variation Modeling for General Time Series Analysis}. In \bibinfo{booktitle}{\emph{International Conference on Learning Representations}}.
\newblock


\bibitem[Xie et~al\mbox{.}(2012)]%
        {www3}
\bibfield{author}{\bibinfo{person}{Sihong Xie}, \bibinfo{person}{Guan Wang}, \bibinfo{person}{Shuyang Lin}, {and} \bibinfo{person}{Philip~S. Yu}.} \bibinfo{year}{2012}\natexlab{}.
\newblock \showarticletitle{Review Spam Detection via Time Series Pattern Discovery}. In \bibinfo{booktitle}{\emph{Proceedings of the 21st International Conference on World Wide Web}} (Lyon, France) \emph{(\bibinfo{series}{WWW '12 Companion})}. \bibinfo{publisher}{Association for Computing Machinery}, \bibinfo{address}{New York, NY, USA}, \bibinfo{pages}{635–636}.
\newblock
\showISBNx{9781450312301}
\urldef\tempurl%
\url{https://doi.org/10.1145/2187980.2188164}
\showDOI{\tempurl}


\bibitem[Xu et~al\mbox{.}(2018)]%
        {donut}
\bibfield{author}{\bibinfo{person}{Haowen Xu}, \bibinfo{person}{Wenxiao Chen}, \bibinfo{person}{Nengwen Zhao}, \bibinfo{person}{Zeyan Li}, \bibinfo{person}{Jiahao Bu}, \bibinfo{person}{Zhihan Li}, \bibinfo{person}{Ying Liu}, \bibinfo{person}{Youjian Zhao}, \bibinfo{person}{Dan Pei}, \bibinfo{person}{Yang Feng}, {et~al\mbox{.}}} \bibinfo{year}{2018}\natexlab{}.
\newblock \showarticletitle{Unsupervised anomaly detection via variational auto-encoder for seasonal kpis in web applications}. In \bibinfo{booktitle}{\emph{Proceedings of the 2018 world wide web conference}}. \bibinfo{pages}{187--196}.
\newblock


\bibitem[Xu et~al\mbox{.}(2022)]%
        {anomaly-transformer}
\bibfield{author}{\bibinfo{person}{Jiehui Xu}, \bibinfo{person}{Haixu Wu}, \bibinfo{person}{Jianmin Wang}, {and} \bibinfo{person}{Mingsheng Long}.} \bibinfo{year}{2022}\natexlab{}.
\newblock \showarticletitle{Anomaly Transformer: Time Series Anomaly Detection with Association Discrepancy}. In \bibinfo{booktitle}{\emph{International Conference on Learning Representations}}.
\newblock
\urldef\tempurl%
\url{https://openreview.net/forum?id=LzQQ89U1qm_}
\showURL{%
\tempurl}


\bibitem[Xu et~al\mbox{.}(2021)]%
        {ctr}
\bibfield{author}{\bibinfo{person}{Zhiqiang Xu}, \bibinfo{person}{Dong Li}, \bibinfo{person}{Weijie Zhao}, \bibinfo{person}{Xing Shen}, \bibinfo{person}{Tianbo Huang}, \bibinfo{person}{Xiaoyun Li}, {and} \bibinfo{person}{Ping Li}.} \bibinfo{year}{2021}\natexlab{}.
\newblock \showarticletitle{Agile and Accurate CTR Prediction Model Training for Massive-Scale Online Advertising Systems}. In \bibinfo{booktitle}{\emph{Proceedings of the 2021 International Conference on Management of Data}} (Virtual Event, China) \emph{(\bibinfo{series}{SIGMOD '21})}. \bibinfo{publisher}{Association for Computing Machinery}, \bibinfo{address}{New York, NY, USA}, \bibinfo{pages}{2404–2409}.
\newblock
\showISBNx{9781450383431}
\urldef\tempurl%
\url{https://doi.org/10.1145/3448016.3457236}
\showDOI{\tempurl}


\bibitem[Zhang et~al\mbox{.}(2022b)]%
        {tfad}
\bibfield{author}{\bibinfo{person}{Chaoli Zhang}, \bibinfo{person}{Tian Zhou}, \bibinfo{person}{Qingsong Wen}, {and} \bibinfo{person}{Liang Sun}.} \bibinfo{year}{2022}\natexlab{b}.
\newblock \showarticletitle{TFAD: A Decomposition Time Series Anomaly Detection Architecture with Time-Frequency Analysis}. In \bibinfo{booktitle}{\emph{Proceedings of the 31st ACM International Conference on Information \& Knowledge Management}}. \bibinfo{pages}{2497--2507}.
\newblock


\bibitem[Zhang et~al\mbox{.}(2022a)]%
        {anotransfer}
\bibfield{author}{\bibinfo{person}{Shenglin Zhang}, \bibinfo{person}{Zhenyu Zhong}, \bibinfo{person}{Dongwen Li}, \bibinfo{person}{Qiliang Fan}, \bibinfo{person}{Yongqian Sun}, \bibinfo{person}{Man Zhu}, \bibinfo{person}{Yuzhi Zhang}, \bibinfo{person}{Dan Pei}, \bibinfo{person}{Jiyan Sun}, \bibinfo{person}{Yinlong Liu}, {et~al\mbox{.}}} \bibinfo{year}{2022}\natexlab{a}.
\newblock \showarticletitle{Efficient kpi anomaly detection through transfer learning for large-scale web services}.
\newblock \bibinfo{journal}{\emph{IEEE Journal on Selected Areas in Communications}} \bibinfo{volume}{40}, \bibinfo{number}{8} (\bibinfo{year}{2022}), \bibinfo{pages}{2440--2455}.
\newblock


\bibitem[Zhang et~al\mbox{.}(2019)]%
        {ATAD}
\bibfield{author}{\bibinfo{person}{Xu Zhang}, \bibinfo{person}{Qingwei Lin}, \bibinfo{person}{Yong Xu}, \bibinfo{person}{Si Qin}, \bibinfo{person}{Hongyu Zhang}, \bibinfo{person}{Bo Qiao}, \bibinfo{person}{Yingnong Dang}, \bibinfo{person}{Xinsheng Yang}, \bibinfo{person}{Qian Cheng}, \bibinfo{person}{Murali Chintalapati}, {et~al\mbox{.}}} \bibinfo{year}{2019}\natexlab{}.
\newblock \showarticletitle{Cross-dataset Time Series Anomaly Detection for Cloud Systems.}. In \bibinfo{booktitle}{\emph{USENIX Annual Technical Conference}}. \bibinfo{pages}{1063--1076}.
\newblock


\bibitem[Zhang et~al\mbox{.}(2005)]%
        {r4}
\bibfield{author}{\bibinfo{person}{Yin Zhang}, \bibinfo{person}{Zihui Ge}, \bibinfo{person}{Albert Greenberg}, {and} \bibinfo{person}{Matthew Roughan}.} \bibinfo{year}{2005}\natexlab{}.
\newblock \showarticletitle{Network anomography}. In \bibinfo{booktitle}{\emph{Proceedings of the 5th ACM SIGCOMM conference on Internet Measurement}}. \bibinfo{pages}{30--30}.
\newblock


\bibitem[Zhao et~al\mbox{.}(2023)]%
        {zhao2023robust}
\bibfield{author}{\bibinfo{person}{Chenyu Zhao}, \bibinfo{person}{Minghua Ma}, \bibinfo{person}{Zhenyu Zhong}, \bibinfo{person}{Shenglin Zhang}, \bibinfo{person}{Zhiyuan Tan}, \bibinfo{person}{Xiao Xiong}, \bibinfo{person}{LuLu Yu}, \bibinfo{person}{Jiayi Feng}, \bibinfo{person}{Yongqian Sun}, \bibinfo{person}{Yuzhi Zhang}, {et~al\mbox{.}}} \bibinfo{year}{2023}\natexlab{}.
\newblock \showarticletitle{Robust Multimodal Failure Detection for Microservice Systems}.
\newblock \bibinfo{journal}{\emph{29th ACM SIGKDD Conference on Knowledge Discovery and Data Mining (KDD)}} (\bibinfo{year}{2023}).
\newblock


\bibitem[Zhao et~al\mbox{.}(2019)]%
        {zhao2019automatic}
\bibfield{author}{\bibinfo{person}{Nengwen Zhao}, \bibinfo{person}{Jing Zhu}, \bibinfo{person}{Yao Wang}, \bibinfo{person}{Minghua Ma}, \bibinfo{person}{Wenchi Zhang}, \bibinfo{person}{Dapeng Liu}, \bibinfo{person}{Ming Zhang}, {and} \bibinfo{person}{Dan Pei}.} \bibinfo{year}{2019}\natexlab{}.
\newblock \showarticletitle{Automatic and generic periodicity adaptation for kpi anomaly detection}.
\newblock \bibinfo{journal}{\emph{IEEE Transactions on Network and Service Management}} \bibinfo{volume}{16}, \bibinfo{number}{3} (\bibinfo{year}{2019}), \bibinfo{pages}{1170--1183}.
\newblock


\bibitem[Zhou et~al\mbox{.}(2021)]%
        {informer}
\bibfield{author}{\bibinfo{person}{Haoyi Zhou}, \bibinfo{person}{Shanghang Zhang}, \bibinfo{person}{Jieqi Peng}, \bibinfo{person}{Shuai Zhang}, \bibinfo{person}{Jianxin Li}, \bibinfo{person}{Hui Xiong}, {and} \bibinfo{person}{Wancai Zhang}.} \bibinfo{year}{2021}\natexlab{}.
\newblock \showarticletitle{Informer: Beyond efficient transformer for long sequence time-series forecasting}. In \bibinfo{booktitle}{\emph{Proceedings of the AAAI conference on artificial intelligence}}. \bibinfo{pages}{11106--11115}.
\newblock


\bibitem[Zhou et~al\mbox{.}(2022)]%
        {fedformer}
\bibfield{author}{\bibinfo{person}{Tian Zhou}, \bibinfo{person}{Ziqing Ma}, \bibinfo{person}{Qingsong Wen}, \bibinfo{person}{Xue Wang}, \bibinfo{person}{Liang Sun}, {and} \bibinfo{person}{Rong Jin}.} \bibinfo{year}{2022}\natexlab{}.
\newblock \showarticletitle{Fedformer: Frequency enhanced decomposed transformer for long-term series forecasting}. In \bibinfo{booktitle}{\emph{International Conference on Machine Learning}}. PMLR, \bibinfo{pages}{27268--27286}.
\newblock


\end{thebibliography}

\end{document}